\pdfoutput=1

\documentclass[11pt]{article}

\usepackage[final]{acl} 
\usepackage[utf8]{inputenc} 
\usepackage[T1]{fontenc}    
\usepackage{url}            
\usepackage{booktabs}       
\usepackage{amsfonts}       
\usepackage{nicefrac}       
\usepackage{microtype}      

\usepackage{times}
\usepackage{latexsym}
\usepackage{tikz}
\usetikzlibrary{shapes,decorations}
\usepackage{amsmath,amssymb, bbm}
\usepackage{pifont}
\usepackage{mathtools}
\usepackage{caption}
\usepackage{subcaption}
\usepackage{tabularx}

\usepackage{graphicx}
\usepackage{multicol}
\usepackage{multirow}
\usepackage{booktabs} 
\def\savelastnode{\pgfextra\edef\tmpA{\tikzlastnode}\endpgfextra}
\def\restorelastnode{\pgfextra\edef\tikzlastnode{\tmpA}\endpgfextra}
\tikzstyle{mybox} = [draw=black, fill=yellow!20, thick,
    rectangle, rounded corners, inner sep=10pt, inner ysep=20pt]
\tikzstyle{mynewbox} = [draw=black, fill=yellow!20, thick,
    rectangle, rounded corners, inner sep=5pt, inner ysep=3pt]
\tikzstyle{fancytitle} =[fill=black, text=white]
\tikzstyle{title} = [append after command={%
    \savelastnode node[fancytitle,right=10pt] at (\tikzlastnode.north west)%
    {\large{#1}}\restorelastnode}]

\newcommand{\g}{\, \mid \,}
\newcommand{\bluebold}[1]{\textcolor{Blue}{\large{\textbf{#1}}}}

\usepackage{soul}

\newcommand{\inconsist}[1]{%
  \begingroup
  \sethlcolor{yellow}%
  \textcolor{red}{\hl{#1}}%
  \endgroup
}

\newcommand{\correct}[1]{%
  \begingroup
  \sethlcolor{pink}%
  \textcolor{red}{\hl{#1}}%
  \endgroup
}

\newcommand{\mixed}[1]{%
  \begingroup
  \sethlcolor{Dandelion}%
  \textcolor{red}{\hl{#1}}%
  \endgroup
}

\usepackage{ulem}

\newcommand{\ours}{\textsc{AutoHallusion}}

\usepackage{booktabs} 
\usepackage{makecell}

\usepackage{afterpage}
\usepackage{floatpag}

\usepackage{times}
\usepackage{latexsym}

\usepackage[T1]{fontenc}

\usepackage[utf8]{inputenc}

\usepackage{microtype}

\usepackage{inconsolata}

\usepackage{graphicx}

\usepackage{hyperref}       
\usepackage{url}            
\usepackage{booktabs}       
\usepackage{amsfonts}       
\usepackage{nicefrac}       
\usepackage{mdwlist}
\usepackage{tikz}
\definecolor{darkyellow}{RGB}{196,146,67}
\newif\ifcomment\commenttrue

\ifcomment

    \newcommand{\wxy}[1]{\textbf{\textcolor{blue}{$\leftarrow$ #1}}}
    \newcommand{\dianqi}[1]{{\color{purple}{\small{\bf [Dianqi: #1]}}}}
\else
\newcommand{\wxy}[1]{}
\newcommand{\dianqi}[1]{}

\fi

\title{\ours{}: Automatic Generation of Hallucination Benchmarks for Vision-Language Models}

\author{
Xiyang Wu\thanks{Equal contribution.} 
~~~Tianrui Guan\footnotemark[1]
~~~Dianqi Li \\
\bf{
~~~Shuaiyi Huang
~~~Xiaoyu Liu
~~~Xijun Wang
~~~Ruiqi Xian 
~~~Abhinav Shrivastava}\\
\bf{
~~~Furong Huang
~~~Jordan Lee Boyd-Graber
~~~Tianyi Zhou 
~~~Dinesh Manocha }
\\ \\
University of Maryland, College Park \\
}

\begin{document}
\maketitle

\begin{abstract}

Large vision-language models (LVLMs) are prone to hallucinations, where certain contextual cues in an image can trigger the language module to produce overconfident and incorrect reasoning about abnormal or hypothetical objects.
While some benchmarks have been developed to investigate LVLM hallucinations, they often rely on hand-crafted corner cases whose failure patterns may not generalize well. Additionally, fine-tuning on these examples could undermine their validity. To address this, we aim to scale up the number of cases through an automated approach, reducing human bias in crafting such corner cases.
This motivates the development of \ours{}, the first automated benchmark generation approach that employs several key strategies to create a diverse range of hallucination examples.
Our generated visual-question pairs pose significant challenges to LVLMs, requiring them to overcome contextual biases and distractions to arrive at correct answers.
\ours{} enables us to create new benchmarks at the minimum cost and thus overcomes the fragility of hand-crafted benchmarks.
It also reveals common failure patterns and reasons, providing key insights to detect, avoid, or control hallucinations. 
Comprehensive evaluations of top-tier LVLMs, e.g., GPT-4V(ision), Gemini Pro Vision, Claude 3, and LLaVA-1.5, show a 97.7\% and 98.7\% success rate of hallucination induction on synthetic and real-world datasets of \ours{}, paving the way for a long battle against hallucinations. The codebase and data can be accessed at \href{https://github.com/wuxiyang1996/AutoHallusion}{https://github.com/wuxiyang1996/AutoHallusion}.
\end{abstract}

\section{Introduction}

Large vision-language models (LVLMs)~\cite{2023GPT4VisionSC, liu2023improved} 
bring powerful tools for content generation~\cite{lian2023llm}, autonomous driving~\cite{chen2023driving}, and robotics~\cite{brohan2023rt, guan2024loczson}. 
However, hallucinations~\cite{Zhang2023SirensSI}, i.e., LVLM-generated responses contain information not present in the visual content, 
raise an alarm and limit LVLMs' applications.
Hallucinations occur when LVLMs' perception and reasoning over-rely on the strong priors of their language modules while ignoring the visual sensory inputs~\cite{guan2024hallusionbench}.
%
%
\begin{figure}[t]
    \centering
    \includegraphics[width=\columnwidth]{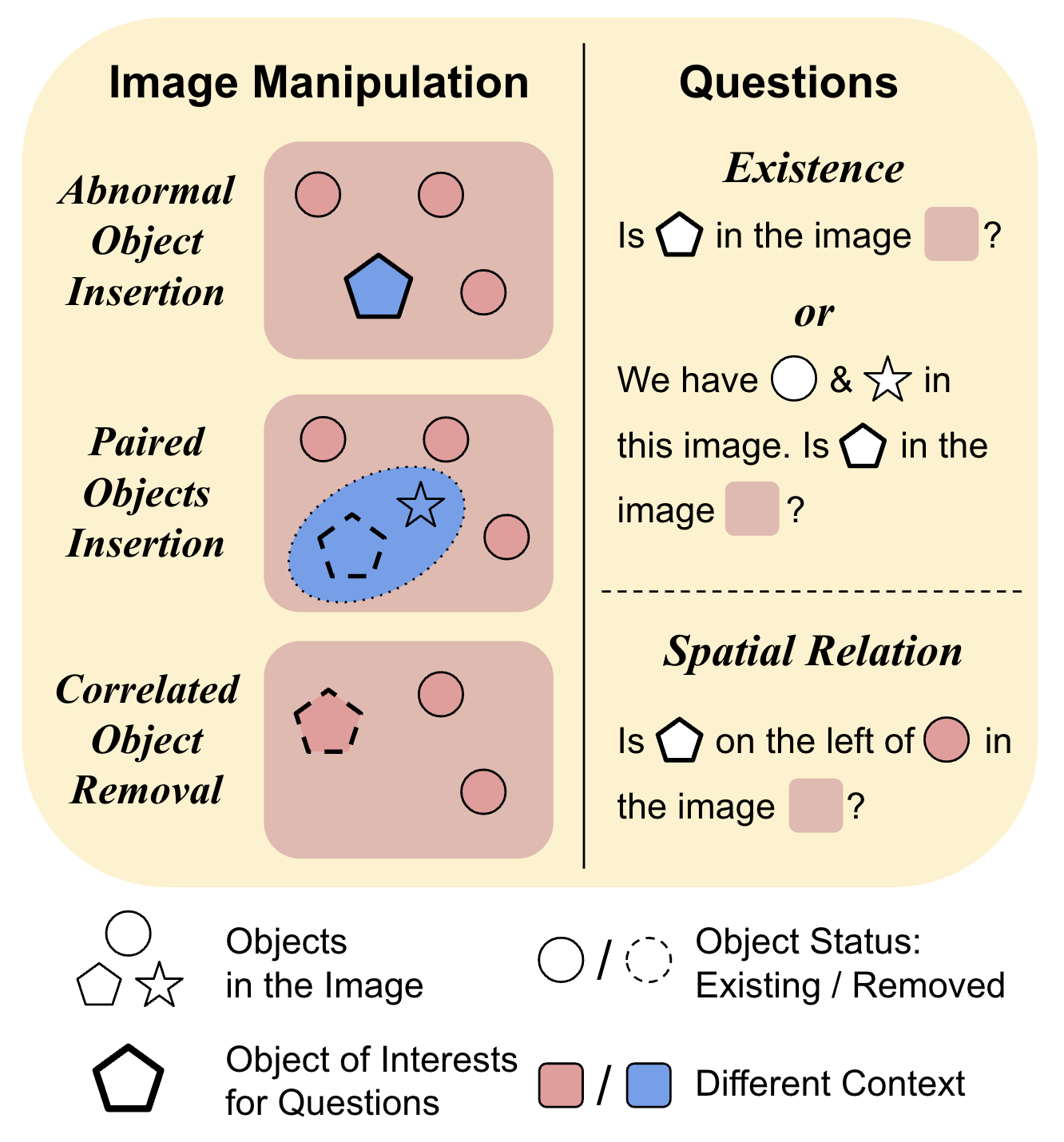}
    \vspace{-8mm}

    \caption{\textbf{\ours{}:} We propose three image manipulation strategies to induce hallucinations: \textit{abnormal object insertion}, \textit{paired object insertion}, and \textit{correlated object removal}, which trigger the conflicts between the images and LVLM priors. Given generated images, we ask LVLMs questions on object existence and their spatial relations for visual question answering.}
    
    \label{fig:cover_pic}
    \vspace{-3mm}
\end{figure}

It is critical for the research community to collect these cases and investigate the reasons behind them. 
With sufficient hallucination examples, finetuning LVLMs on them with the original training data may reduce hallucinations and alleviate those biases. 
However, crafting those cases in previous work requires expensive human labor and is highly time-consuming~\cite{jiang2024hal,rohrbach2018object,li2023evaluating,han2024instinctive,guan2024hallusionbench}. Moreover, it is unclear whether those hand-crafted examples are rare corner cases or indicate general fail patterns. Without an in-depth understanding of the common mechanism generating them, it is hard to extract useful insights to improve LVLMs. 
On the other hand, finetuning on those small benchmarks without sufficient representative examples may lead to overfitting.

To address those challenges, we develop an automated pipeline, \ours{}, to generate diverse hallucination cases and mass-produce them at the minimum cost of human efforts. To generate (image, question) pairs that can trigger the hallucinations of LVLMs, we take a reverse-engineering path: It starts from exploring output answers due to hallucinations, by probing LVLMs' language modules to allocate the strong language priors on certain objects or their contextual relations. It then creates (1) an image containing objects that contradict the probed priors (the presumed answers), and (2) questions on two types of conflicts, the existence of contextual-related objects and their spatial relationships. 
If the LVLM reasoning is biased or dominated by the language prior, it tends to generate incorrect or inconsistent responses conflicting with the ground truth in the images, hence the hallucinations. We provide an optimization formulation and develop three principal strategies, \textit{abnormal object insertion}, \textit{paired object insertion}, and \textit{correlated object removal}, to manipulate the objects in a scene and thus create images conflicting with the language prior, as illustrated in Figure \ref{fig:cover_pic}. \looseness-1

The detailed designs of these hallucination strategies are inspired by \textit{schema}~\cite{dimaggio1997culture, boutyline2021cultural, rumelhart2017schemata} from cognitive science.
\textit{Schema} refers to the tendency of humans to organize information and interpret the world based on patterns of past experiences\footnote{For example, it is much more common to see a keyboard and a mouse in front of a monitor rather than an octopus.}. Following its concept, \textit{irregular schema} with \textit{cognitive dissonance}~\cite{aronson1969theory, harmon2019introduction}, e.g., an octopus in front of a monitor, and \textit{breaking a schema} with \textit{expectancy violation}~\cite{burgoon1993interpersonal, burgoon1988nonverbal}, e.g., the absence of a keyboard and a mouse in front of a monitor, can both induce contradictions and discomforts in the memory. The three strategies reveal common patterns and mechanisms of how hallucinations are generated, hence providing critical insights to detect, combat, avoid, or control hallucinations of LVLMs.

\paragraph{Main contributions:} 
Inspired by an analogy to human cognition in terms of \textit{schema}, we investigate the mechanism of hallucinations in LVLMs by reverse-engineering (image, question) pairs 
with probed language priors and biases. We develop \ours{} that synthesizes images by manipulating the objects in the scenes to conflict with LVLMs' memory (i.e., its language priors), and generates questions about the conflicts. 
The novelties of our work can be summarized as:
\begin{itemize*}
    \item We propose the first automatic generation approach of hallucination benchmarks, with a high-level formulation and three principal strategies, inspired by {\it schema} in cognitive science, to trigger LVLM hallucinations.  
    \item We develop novel probing methods to extract and investigate the contextual biases in the language priors that cause hallucinations. We further introduce two evaluation metrics to detect hallucinations. 
    \item We evaluate SOTA LVLMs, including GPT-4V(ision), Gemini Pro Vision, Claude 3, and LLaVA-1.5, on benchmarks by \ours{}.
    Our method achieves success rates of 97.7\% and 98.7\% in inducing LVLM hallucinations on synthetic and real-world data, respectively. Based on our experiments, we curated a benchmark dataset to further evaluate model performance.
    
\end{itemize*}


\section{Related Work}



\paragraph{Vision-Language Models (VLMs).}
The recent increase in large language models (LLMs)~\cite{liu2024largelanguagemodelscausal}, including GPT-3~\cite{brown2020language}, PaLM~\cite{chowdhery2023palm}, and BLOOM~\cite{le2023bloom}, has significantly improved natural language processing. LLaMA~\cite{touvron2023llama} further advanced this field, and models like Alpaca~\cite{taori2023stanford}, inspired by InstructGPT~\cite{ouyang2022training} and ChatGPT, utilized human-annotated data to refine LLaMA, enhancing its interaction abilities. Additionally, Large Visual Language Models (LVLMs) such as GPT-4~\cite{achiam2023gpt}, Flamingo~\cite{alayrac2022flamingo}, Bard~\cite{bard2023}, MiniGPT-4~\cite{zhu2023minigpt}, InstructBLIP~\cite{dai2024instructblip}, and LLaVA~\cite{liu2024visual} have developed. These models combine visual and language processing to manage both textual and visual inputs and produce textual outputs. Their architecture generally includes a visual encoder (often based on CLIP~\cite{radford2021learning}), a modality connection module, and an LLM. LVLMs excel in generating text descriptions from images and multi-modal learning through pre-training on image-text pairs~\cite{liu2024survey} and instruction-tuning with various tasks~\cite{liu2023aligning, ouyang2022training, xu2023wizardlm}. However, addressing hallucinations in their textual outputs remains a challenge, emphasizing the need for reliability and accuracy in real-world applications.

\paragraph{Benchmarks.}
Several benchmarks have been developed to assess hallucination in VLMs in various aspects. CHAIR~\cite{rohrbach2018object} evaluates object hallucination by measuring word accuracy against ground-truth sentences and segmentation for 80 MSCOCO objects. POPE~\cite{li2023evaluating} improves upon CHAIR for better stability and flexibility while OpenCHAIR~\cite{ben2023mocha} extends CHAIR to open-vocabulary settings. HallusionBench~\cite{guan2024hallusionbench} targets visual commonsense and reasoning with 455 visual-question control pairs. Hal-Eval~\cite{jiang2024hal} introduces and focuses on event hallucination while CorrelationQA~\cite{han2024instinctive} examines the impact of spurious visual inputs. Our work differs from previous benchmarks~\cite{qian2024easy, wang2024haloquest} using hand-crafted examples from datasets like COCO~\cite{lin2014microsoft} or Visual Genome~\cite{krishna2017visual} by employing an auto-generated approach to synthesize hallucination cases through contextual influences, making our method more effective and scalable.

\paragraph{Object Hallucination.}
Large Vision Language Models (LVLMs) hold great potential but struggle with object hallucination, generating incorrect descriptions that include nonexistent objects or omit key details. This problem can adversely affect applications in robotics~\cite{wu2024safety,liu2023llm}, medical imaging~\cite{wang2023chatcad,hu2023advancing}, and human-computer interaction~\cite{brie2023evaluating}. Object hallucination in LVLMs manifests as fictional objects, false attributes, or inaccurate relationships between objects~\cite{gunjal2024detecting, zhai2023halle}. Previous methods, like fine-tuning smaller multimodal models~\cite{biten2022let,kim2023exposing}, are less effective for LVLMs due to their distinct architectures. Recent efforts focus on improving dataset quality for fine-tuning~\cite{li2023m,liu2023aligning}, but acquiring such data remains labor-intensive. Metrics like CHAIR~\cite{rohrbach2018object} and POPE~\cite{li2023evaluating}, which assess caption relevance and hallucination levels, are crucial for evaluation. Standard text quality metrics can be misleading, as high scores may still correlate with significant hallucination. In this paper, we investigates contextual biases in language priors causing hallucinations and introduces two new metrics for more effective detection.

\begin{figure*}[t]
    \centering
    \includegraphics[width=\textwidth]{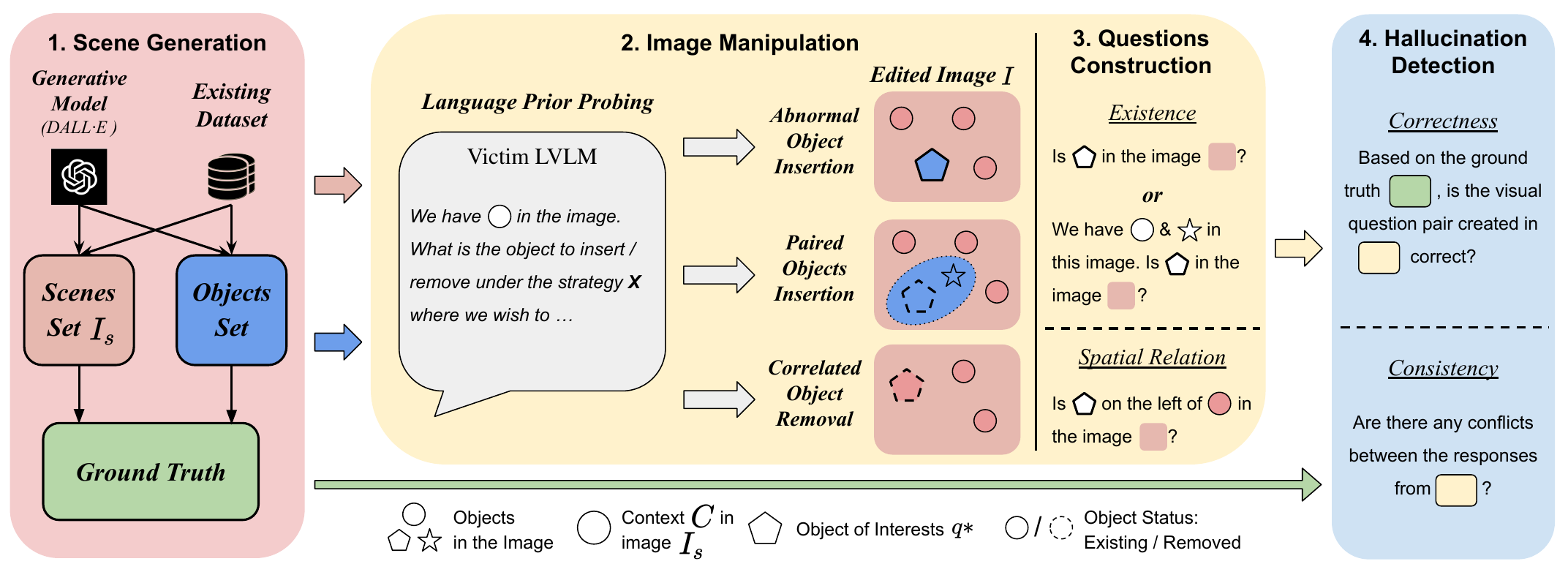}

     \caption{\textbf{Overview of \ours{}.} We first automatically generate the scenes set and objects set (pink). After that, we use text to probe the language prior of the victim LVLM and then propose three manipulation strategies to induce hallucination in scene images(yellow). We finally use two metrics to detect hallucinations (blue).
    }

    \label{fig:teaser}
\vspace{-3mm}
\end{figure*}

\section{Problem Formulation}
\label{Section: Problem Formulation}

Pronounced bias in LLMs hinders the reasoning capability of LVLMs, resulting in hallucinations from the images~\cite{guan2024hallusionbench}. 
Inspired by this, we target the biases in LLMs to induce hallucinations in LVLMs. 




\noindent\textbf{Definitions and Objective.} Our objective is to find things that are correlated in the LLM but not present in the picture to induce hallucinations in LVLMs. Let $f_\mathtt{LVLM}$, $f_\mathtt{LLM}$ denote the LVLM and its LLM component, respectively. $f_\mathtt{LVLM}(\textit{image}, \textit{query})$ can take an image-query pair as inputs, and $f_\mathtt{LLM}(\textit{context}, \textit{query})$ can take a text-only context-query pair as inputs. We use sets as universal representations for the images and texts and detailed as below. 

We denote $\mathcal{V}$ as the set of all contextual elements in an image $I$, where each element can be an object, an attribute associated with an object, or the relation between/among multiple objects, etc.\footnote{similar to the visual genome dataset.}
%
%
These elements in the set can be considered as a statement, which could be either affirmative or negative. Similarly, for text modality, we denote $\mathcal{Q}$ as the set containing objects of interest for questions and $\mathcal{C}$ as the set of objects in this scene for context.
We use a mapping function $\mathit{T}(\cdot)$ to transform a set of contextual elements into a text, which can be either a description from $\mathcal{C}$ or a query question from $\mathcal{Q}$.

Finally, we introduce the contextual distance $d[\cdot,\cdot]$ between two descriptions or texts. When two pieces of text convey similar information or affirm each other, the contextual distance $d$ is considered small; otherwise, the contextual distance is large. Let $y_{\mathcal{Q}}$ be the ground truth answer set with respect to the query set $\mathcal{Q}$ given the image $I$. The objective function can be formulated as follows:\vspace{-2mm}
\begin{align}
    \max\limits_{I,\mathcal Q,\mathcal C} &~~~~d[f_\mathtt{LVLM}(I, \mathit{T}({\mathcal Q})), y_{Q}] \label{equ:f2s} \\[-1mm]
    s.t. & ~~~~d[f_\mathtt{LVLM}(I, \mathit{T}({\mathcal Q})), f_\mathtt{LLM}(\mathit{T}({\mathcal C}), \mathit{T}({\mathcal Q}))]\leq \epsilon, \nonumber \\
    & ~~~~\mathcal C\subseteq \mathcal V, \mathcal Q\cap \mathcal C=\emptyset. \label{equ:f2constrain}
\end{align}


The objective function~\eqref{equ:f2s} maximizes the distance between the generated text $f_\mathtt{LVLM}$ and $y_{\mathcal{Q}}$ to produce hallucination. To leverage and probe the bias in the language components $f_\mathtt{LLM}$ of the victim LVLM, we use constrain~\eqref{equ:f2constrain} to control the discrepancies between responses, with and without visual input, within a tolerance $\epsilon$. This ensures that the answer is dominated by the prior language component rather than the visual component. The visual information $\mathcal{V}$ from the image $I$ provided to $f_\mathtt{LVLM}$ is partially converted to a text, $\mathit{T}({\mathcal C})$, as the input to $f_\mathtt{LLM}$, therefore $\mathcal C\subseteq \mathcal V$.

\noindent\textbf{Remark.} It is important for the language component $f_\mathtt{LLM}$ to have the constraint $\mathcal Q\cap \mathcal C=\emptyset$. If the interested element from $\mathcal Q$ is directly given in the context $\mathcal C$, it would be difficult to exploit the bias of $f_\mathtt{LLM}$ since it is mostly likely to answer based on the provided context $\mathcal C$. For example, if a fact is directly given in the prompt, it is hard for the model to make a contradictory claim. In addition, $\mathcal Q$ is not required to be the subset of $\mathcal V$ since we can ask questions on objects that are not included in the image $I$.

\noindent\textbf{Approach.}
It is hard to optimize $I$, $\mathcal Q$, and $\mathcal C$ by directly optimizing Eq.~\eqref{equ:f2s}. 
%
Instead, we probe the LVLM and the language prior from its LLM component to determine $(\mathcal Q, \mathcal C)$ such that the elements in $\mathcal Q$ are highly likely (or unlikely, depending on the attack strategy) to be present with $\mathcal C$ in the same scene. 
Such bias in the language prior helps us achieve the constraint~\eqref{equ:f2constrain}.
This ensures that the language prior is strong and highly confident on the co-occurrence of $(\mathcal Q, \mathcal C)$, i.e., $\Pr(\mathcal Q \g \mathcal C)\leq \delta$ ($\mathcal Q$ is abnormal given $\mathcal C$) or $\Pr(\mathcal Q \g \mathcal C)\geq 1-\delta$ ($\mathcal Q$ is hypothetical given $\mathcal C$), where $\Pr(\mathcal Q \g \mathcal C)$ is the probability of the existence of elements in $\mathcal Q$ given the presence of $\mathcal C$ and $\delta$ is the confidence level. If the assumption on $(\mathcal Q, \mathcal C)$ pairs that the LVLM reasoning is dominated by its language prior, \textit{i.e.} Eq.~\eqref{equ:f2s} holds true, we can create $I$ from such $(\mathcal Q, \mathcal C)$ pairs to maximize the discrepancy in Eq.~\eqref{equ:f2s}.
\section{Methodology}
\label{Methodology}

The overall pipeline of our methodology is presented in Fig. \ref{fig:teaser}. 
We break down the automated procedure of creating hallucination cases into 4 stages: scene generation, image manipulation, question construction, and hallucination detection. 
Questions constructed to induce potential hallucination cases vary depending on these strategies, mainly focusing on object existence and spatial relations between the target object and others. We detect hallucinations through correctness and consistency among answers generated by the victim LVLM.


\subsection{Scene Generation}
\label{Methodology: Scene Generation}
First, we want to create a scene image $I_s$ with a strong context $\mathcal{C}$ so that it would be easier to extract bias and incur hallucination.
Given a random scene name or a brief description, we use the target LVLM to generate and expand on the contextual elements $\mathcal{C}$ within the scene. With these descriptions and details, we employ a diffusion model or an image generation model like DALL-E-3~\cite{dalle3} to create an image $I_s$ rich in context, incorporating the provided scene information and relevant objects that are listed in the context $\mathcal{C}$. Alternatively, $I_s$ can be obtained from an existing dataset, assuming the images are coherent, natural, and contain several correlated elements. We use Owl-ViT~\cite{minderer2022simple} to ground the contextual elements of $I_s$ and verify the context $\mathcal{C}$.



\subsection{Image Manipulation}
\label{Methodology: Manipulation of Target Object}

Once we have a scene image $I_s$ rich in context, we want to use $\mathcal{C}$ to probe the LLM component $f_\mathtt{LLM}$ of the victim model and find a target object, which is used to modify $I_s$. This target object is not only used to manipulate $I_s$, but also used to construct the questions $\mathcal{Q}$. Once we find a suitable $\mathcal{Q}$ based on $\mathcal{C}$, we can modify $I_s$ and manipulate the target object to obtain the final $I$. 

Our hallucination attack focuses one contextual element $q^*$ retrieved from the query set $\mathcal Q$. Since $\mathcal Q$ is not bounded to all the visual elements $\mathcal V$ from the image, the modification can be either object insertion or removal. Our manipulation strategies are explained as follows:

\subsubsection{Abnormal Object Insertion}
The abnormal object insertion strategy tries to insert an object not related to the existing contextual elements into the scene image $I_s$. 
For example, given an image of an office scene, a suitable abnormal object that contradicts this context could be a cooking pot. 

The query question $q^*$, which is also the abnormal object for insertion, should have the \textbf{maximum sum of distances} between its language prior and the ground truth information across all contextual elements in $\mathcal C$. We bound the retrieval process as:
\begin{align}
    q^* = \arg\max_{q \in \mathcal Q} \sum_{c \in \mathcal C} d(f_\mathtt{LLM}(\mathit{T}(c), \mathit{T}(q)), y_{q})
\end{align}

In practice, we use DALL-E 3~\cite{dalle3} to create this image for the abnormal object or choose the object's image from an existing database. To guarantee the insertion is successful and does not introduce any artifacts, we simply stitch the object into the scene image after removing the background of the object image~\cite{backgroundremover2021} instead of using diffusion or an in-painting method.

\subsubsection{Paired Object Insertion}
The paired object insertion strategy uses target LVLM to determine the paired objects with a strong correlation, like coffee makers and coffee beans. In this strategy, we insert only one of two objects from the pair and ask questions about the other. 

We formulate this image manipulation process into finding the query question $q^*$ with the \textbf{minimum element-wise distance} between its language prior and the ground truth information among all contextual elements in $\mathcal C$:
\begin{align}
    q^* = \arg\min_{q \in \mathcal Q} \min_{c \in \mathcal C} d(f_\mathtt{LLM}(\mathit{T}(c), \mathit{T}(q)), y_{q})
\end{align}

\subsubsection{Correlated Object Removal}
The correlated object removal strategy removes the existing object from the generated scene image $I_s$, while the removed object has a strong correlation with multiple contextual elements within $I_s$. We use Owl-ViT~\cite{minderer2022simple} to extract contextual elements $\mathcal{C}$ within the image and choose the object from the detected object list as the adversary object $q^*$ to remove using the in-painting function from Dall-E-2~\cite{dalle3}.

We choose the adversary object $q^*$ by searching for the object with the \textbf{minimum sum of distances} between its language prior and the ground truth information across all contextual elements in $\mathcal C$:
\begin{align}
    q^* = \arg\min_{q \in \mathcal Q} \sum_{c \in \mathcal C} d(f_\mathtt{LLM}(\mathit{T}(c), \mathit{T}(q)), y_{q})
\end{align}


\noindent\textbf{Intuition.} The purpose of two types of insertion is to add an abnormal element that the model will ignore given the strong context and to insert one of a correlated object pair so the model will hallucinate about the other, respectively. For the removal strategy, we aim to identify and erase the object that has the strongest correlation with the context or the shortest sum of contextual distances to the ground truth of the scene given the query.

\subsection{Question Construction}
\label{Methodology: Question Construction}

We mainly consider two types of questions: the existence of the object and the spatial relation between the objects.


In the existence questions, we ask whether the target object $q^*$ is present in the image. These questions are repeated multiple times, with varying levels of details mentioned in the prompt. For example, we ask the same victim model to generate an image caption and add this text in front of the query question because such supplementary information may serve as another source of language prior that misleads the victim model. 
In addition, we also ask existence questions on objects that are missing in the image caption generated by the victim model because it has a higher probability of missing this object again in the existence question. 


In the spatial relation question, we ask about the relative positions of the target object and the scene objects. Given the bounding boxes, it's easy to obtain the following spatial relations: \textit{Left}, \textit{Right}, \textit{Above}, \textit{Below}, \textit{Front} (when the perturbed object overlaps with the scene object).
To address issues made by overlapping boxes, we remove new ones with large overlaps, retaining only the highest confidence bounding box per object. Spatial relations are determined by comparing object center positions, ensuring no overlaps and effective relation extraction.
%
Spatial relation questions are asked with multiple levels of contextual information from the image, including vanilla (no extra information), single and concatenated object-level description, and the detailed caption for the whole image, all of which are generated by the victim model.

\subsection{Hallucination Detection}
\label{Methodology: Hallucination Detection}
We use GPT-4V-Turbo~\cite{2023GPT4VisionSC} to evaluate the correctness of the predicted answer by the victim model and the ground truth. There are two criteria to determine whether hallucination occurs with different levels of reliability:

\begin{enumerate*}
    \item \textbf{Correctness:} Since we know the ground truth existence and relations of objects, we can easily determine the correctness of the visual question pairs. This criterion is the most straightforward, but it does not account for any generation errors or background-removal artifacts from the pipeline. If some of the steps fail, the ground truth may not be reliable. 
    \item \textbf{Consistency:} In this criterion, we want to consider the consistency of the model outputs, which does not rely on whether the ground truth is accurate. For example, if we ask about the existence of an object and get different responses, we are certain that one of the responses is hallucinating. We divide the inconsistency hallucination into two categories: \textbf{(1) Response Conflict} happens when LVLMs fail to give consistent answers to questions with different levels of supplementary information provided, and \textbf{(2) Local-Global Conflict} occurs when LVLMs fail to provide answers about the object of interest (local) that are consistent with the caption describing the image related to that object.
\end{enumerate*}

We manually inspect the image-query-ground truth benchmark evaluation results by LVLMs and our inspection confirmed that, among all successful cases, 92.6\% of all probing questions and their corresponding answers are correctly evaluated.

\section{Evaluation and Metrics}


\begin{figure}[t]
    \centering
    \includegraphics[width=0.9\columnwidth]{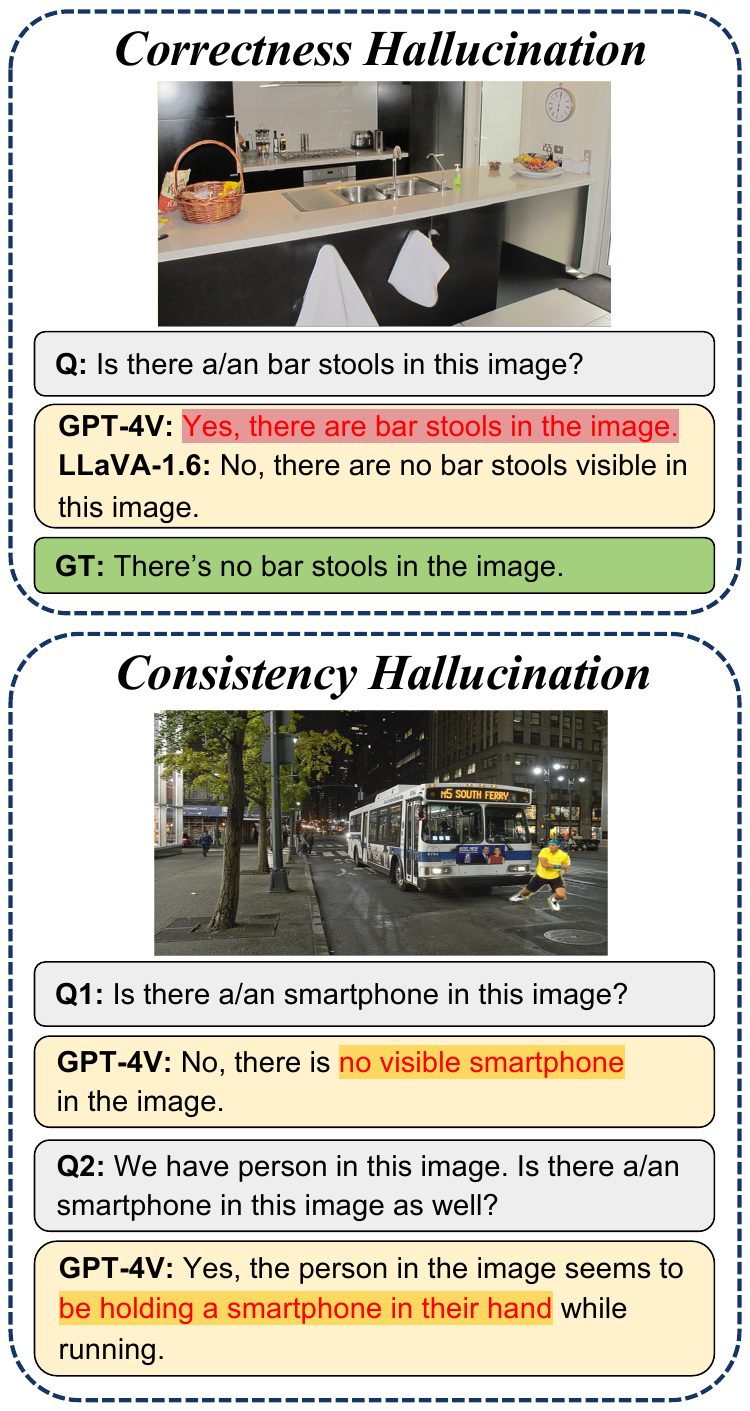}

    \caption{\textbf{Hallucination Cases created by \ours{}.} We highlight hallucination context made by \correct{\textit{Correctness}} and  \inconsist{\textit{Inconsistency}}}

    \label{fig:showcase_main}
\vspace{-5mm}
\end{figure}

\subsection{Implementation Details}
\noindent\textbf{Data Preparation.}
To obtain all the scene images and object images for insertion, we either generated those images with image generation models like DALL-E-3~\cite{dalle3}, or use existing datasets. For image generation, we first use LVLM to fill in more details of the scene with objects for better generation results.
For real-world data, we use the validation dataset from the Common Objects in Context (COCO) dataset~\cite{lin2014microsoft}. We randomly select $126$ samples with sufficient contextual elements provided in the image and around $5,000$ object images segmented from raw images. We edit the scene image by inserting objects retrieved from the database, thinking about the correlated object for the given object, or removing them from the scene.

\begin{table*}[h]
  \begin{center}
  \resizebox{\textwidth}{!}{
    \begin{tabular}{clcccccccccc}
    \toprule
        \multicolumn{2}{c}{} & \multicolumn{5}{c}{\makecell{Synthetic Data}}
        & \multicolumn{5}{c}{\makecell{Real-World Data}} \\
        \cmidrule(lr){3-7} \cmidrule(lr){8-12} 
        \makecell{Manipulation \\ Strategies} & LVLMs &  \makecell{Overall \\ ASR} & \makecell{Overall \\ MASR} & \makecell{Overall \\ CASR} & \makecell{Exi. \\ ASR} & \makecell{Sp. \\ ASR} & \makecell{Overall \\ ASR} & \makecell{Overall \\ MASR} & \makecell{Overall \\ CASR} & \makecell{Exi. \\ ASR} & \makecell{Sp. \\ ASR}\\
        \midrule
        \multirow{5}*{\makecell{Abnormal Obj. \\ Insertion}} & GPT-4V-Turbo \cite{yang2023dawn} & 96.0 & 80.0 & 92.5 & 93.0 & 78.1 & 100.0 & 98.4 & 98.4 & 97.6 & 97.5  \\
        ~ & Gemini Pro Vision \cite{google_gemini} & 97.0 & 90.5 & 90.0 & 84.5 & 89.1 & 100.0 & 100.0 & 97.6 & 97.6 &  94.3 \\
        ~ & Claude 3 \cite{claude3} & 97.4 & 90.7 & 96.0 & 95.3 & 92.3 & 100.0 & 100.0 & 100.0 & 100.0 & 98.4 \\
        ~ & LLaVA-1.5 \cite{liu2023improved} & 97.7 & 94.2 & 94.0 & 97.9 & 96.2 & 100.0 & 100.0 & 98.9 & 98.6 & 95.9 \\
        ~ & miniGPT4 \cite{zhu2023minigpt} & 98.1 & 95.1 & 98.0 & 98.1 & 97.1 & 100.0 & 100.0 & 97.9 & 98.0 & 96.1 \\
        \midrule
        \multirow{5}*{\makecell{Paired Obj. \\ Insertion}} & GPT-4V-Turbo \cite{yang2023dawn} & 99.5 & 93.5 & 97.0 & 91.5 & 81.7 & 100.0 & 100.0 & 99.2 & 99.2 & 100.0\\
        ~ & Gemini Pro Vision \cite{google_gemini} & 100.0 & 100.0 & 99.5 & 99.5 & 85.7 & 100.0 & 99.2 & 100.0 & 99.2 & 90.4\\
        ~ & Claude 3 \cite{claude3} & 100.0 & 99.0 & 99.0 & 99.0 & 95.5 & 100.0 & 99.2 & 100.0 & 97.6 & 99.2\\
        ~ & LLaVA-1.5 \cite{liu2023improved} &  99.7 & 95.1 & 98.9 & 97.6 & 81.8 & 99.7 & 98.5 & 99.3 & 94.5 & 97.8\\
        ~ & miniGPT4 \cite{zhu2023minigpt} & 100.0 & 99.8 & 100.0 & 99.1 & 83.9 & 100.0 & 100.0 & 99.5 & 99.5 & 99.8 \\
        \midrule
        \multirow{5}*{\makecell{Correlated Obj. \\ Removal}} & GPT-4V Turbo \cite{yang2023dawn} & 93.0 & 84.0 & 84.0 & 69.5 & 85.5 & 94.4 & 88.0 & 84.0 & 75.2 & 85.4 \\
        ~ & Gemini Pro Vision\cite{google_gemini} & 95.0 & 92.0 & 93.0 & 77.0 & 91.1 & 96.8 & 95.2 & 92.0 & 77.6 &  94.2 \\
        ~ & Claude 3\cite{claude3} & 99.0 & 98.0 & 89.0 & 92.0 & 88.5 & 98.4 & 98.4 & 94.4 & 96.0 & 89.6 \\
        ~ & LLaVA-1.5\cite{liu2023improved} & 97.1 & 88.9 & 87.4 & 70.8 &  87.4 & 93.1 & 97.6 & 94.6 & 78.1 & 95.7\\
        ~ & miniGPT4 \cite{zhu2023minigpt} & 96.7 & 90.1 & 91.5 & 72.9 & 86.7 & 97.8 & 96.3 & 89.1 & 76.9 &  87.8\\
    \hline
    \end{tabular}
    }
    \end{center}

    \caption{\textbf{Evaluation results of SOTA LVLMs with our \ours{} on synthetic and real-world data.} Our proposed three manipulation strategies achieved high success rates (the higher the better) on synthetic and real-world data.}

    
    \label{main_results}
    \vspace{-3mm}
\end{table*}

In Appendix~\ref{Appendix: Showcase}, we provide showcases for both data preparation methods, including the initial scene and object images and those images after manipulation, which are either generated by DALL-E-2~\cite{dalle3} or queried from the real-world image dataset.

\noindent\textbf{Victim LVLMs.}
We conduct extensive experiments using the proposed \ours{} across the following models: GPT-4V-Turbo~\cite{yang2023dawn}, LLaVA-1.5~\cite{liu2023improved}, Claude 3~\cite{claude3}, Gemini Pro Vision~\cite{google_gemini}, and miniGPT4~\cite{zhu2023minigpt}. 

\noindent\textbf{Implementation Details.} We generate $200$ cases for each experiment. By default, all scene and edited images are $1024 \times 1024$ and inserted objects are $200 \times 200$ for synthetic data. For real-world dataset, we loop over all scene images with proper resizing to fit the input of image models.

For a given generated scene image $I_s$, we use the object detection model~\cite{minderer2022simple} to detect and segment all candidate contextual elements for removal from the image. 
We use the generative image model DALL-E-2 \cite{ramesh2022hierarchical} to in-paint the chosen object for removal. 


\subsection{Evaluation Metrics}

Apart from the overall \textbf{Attack Success Rate (ASR)} of each evaluation category, we mainly use the following evaluation metrics to determine whether hallucination generation is successful: 

\noindent\textbf{Manipulation Attack Success Rate (MASR):} We compare the generated response with the ground truth generated based on the intention of the image generation and editing. However, it is possible that the ground truth of the image is not accurate due to failure during image generation and editing.

\noindent\textbf{Conflict Attack Success Rate (CASR):} We ask a set of questions and try to find conflicts among all responses to those visual questions. Such inconsistency will guarantee that one of the conflicting responses must have been hallucinated and provided an incorrect answer.

\subsection{Main Results}
\label{Sec: Main Results}
Table~\ref{main_results} summarizes the performance of victim LVLMs under our three attack strategies using synthetic and real-world datasets. We achieve high ARS with all three proposed attack strategies in both datasets, showing the effectiveness of our approach to induce hallucinations.


We have the following key observations: (1) Strategies probing inserted objects (Abnormal Object and Paired Object Insertion), achieve higher hallucination attack success rates than those probing absent objects (Correlated Object Removal strategy); (2) Questions probing the existence of objects are more effective to cause hallucinations than questions probing spatial relations; (3) GPT-4V-Turbo is the most robust to hallucination attacks among all victim LVLMs; (4) Our method achieved even higher attack success rates across all LVLMs in the real-world dataset than synthetic data. 
We hypothesize this comes from LVLMs lack of ability to address the complexity and diversity within the real-world data, which causes its higher vulnerability to our attack strategies when using real-world data.
%
For more experimental results, please refer to Appendix~\ref{appendix:exp}.

\subsection{Ablation Studies}
\label{Sec: Ablation Studies}
\paragraph{Object Sizes.}

Table~\ref{ablation_window} shows results for different object sizes from $100 \times 100$ to $400 \times 400$ using an abnormal object insertion strategy with GPT-4V-Turbo, while \ours{} generally uses $200 \times 200$. The findings indicate that larger objects reduce hallucinations, including those from image manipulation and response conflicts. Similar patterns are evident in questions probing existence and spatial relationships. LVLMs are more vulnerable to smaller perturbed objects, as they struggle to encode small images into tokens. However, we attribute this phenomenon comes from visual illusions made by the failure of visual encoders of LVLMs, instead of hallucinations targeting the reasoning abilities of LVLMs.
%
We selected the current object size to balance hallucination attack performance with the reduction of visual illusions. 


\begin{table}[t]
  \begin{center}
  \resizebox{\linewidth}{!}{
    \begin{tabular}{lccccccccccccccccc}
    \toprule
        \multicolumn{1}{c}{} & \multicolumn{3}{c}{\makecell{Overall}}
        & \multicolumn{3}{c}{\makecell{Existence}} & \multicolumn{3}{c}{\makecell{Spatial Relation}} \\
        \cmidrule(lr){2-4} \cmidrule(lr){5-7} \cmidrule(lr){8-10}
         Obj. Size & \makecell{Overall \\ ASR} & \makecell{Overall \\ MASR} & \makecell{Overall \\ CASR} & \makecell{Exi. \\ ASR} & \makecell{Exi. \\ MASR} & \makecell{Exi. \\ CASR} & \makecell{Sp. \\ ASR} & \makecell{Sp. \\ MASR} & \makecell{Sp. \\ CASR}\\
        \midrule
        $100 \times 100$ & 98.0 & 90.0 & 97.5 & 97.0 & 78.5 & 96.0 & 87.5 & 80.6 & 70.0\\
        \textbf{200 $\times$ 200} & 96.0 & 80.0 & 92.5 & 93.0 & 62.0 & 88.5 & 78.1 & 71.2 & 60.6\\
        $300 \times 300$ & 93.5 & 75.0 & 85.5 & 87.0 & 54.0 & 80.5 & 76.3 & 69.4 & 45.0\\
        $400 \times 400$ & 89.5 & 68.5 & 79.0 & 81.0 & 43.5 & 74.0 & 65.6 & 53.8 & 41.9\\
    \hline
    \end{tabular}
    }
    \end{center}
    \vspace{-3mm}
    \caption{\textbf{Ablation on the size of the objects with abnormal object insertion using GPT-4V-Turbo.}} 
    \label{ablation_window}
    \vspace{-3mm}
\end{table}

\begin{figure}[t]
    \centering
    \includegraphics[width=\columnwidth]{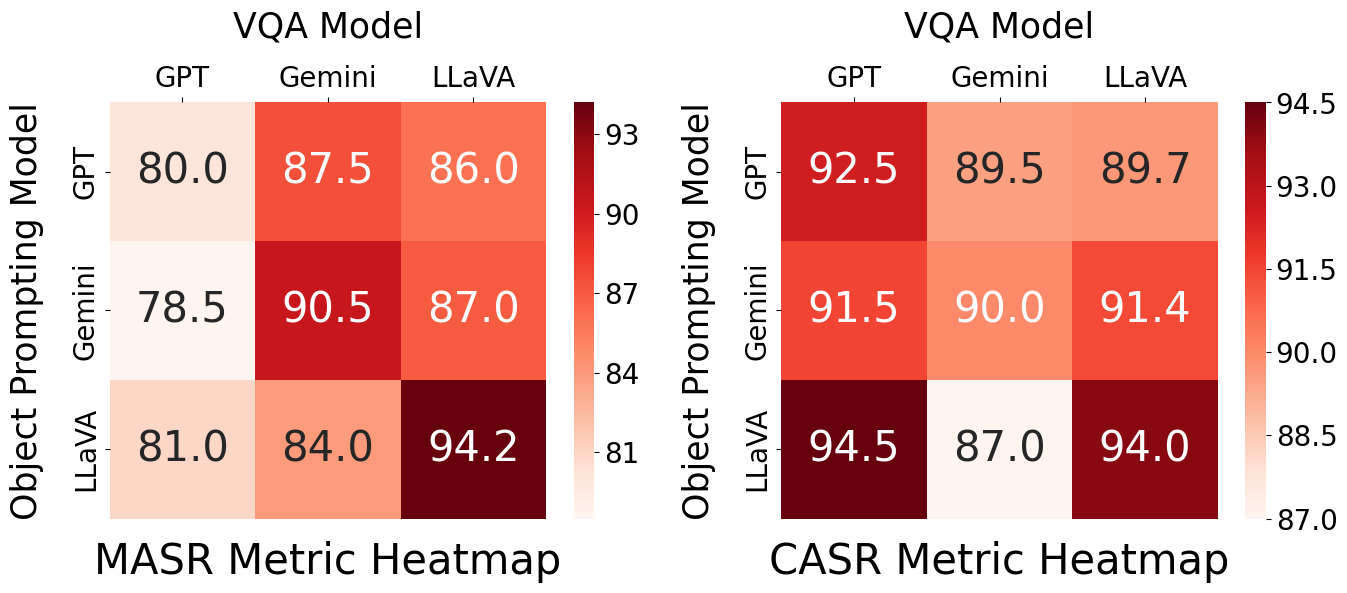}
     \caption{\textbf{Ablation on using different LVLMs for object prompting and VQA tasks.}}
    \label{fig:heatmap}
\vspace{-5mm}
\end{figure}

\paragraph{Object Prompting and VQA Alignment.}
As we mentioned in Section~\ref{Methodology: Manipulation of Target Object} and~\ref{Methodology: Question Construction}, we use the same victim model to prompt objects for image manipulation and perform VQA tasks with constructed questions, which may introduce inherited biases. We conduct ablation experiments to de-bias and evaluate models' performance on each sub-task separately by swapping models for object prompting and VQA with abnormal object retrieval strategy. Fig.~\ref{fig:heatmap} shows the results using different models among GPT-4V-Turbo, Gemini Pro Vision, and LLaVA-1.5 performing abnormal object prompting and VQA tasks. Results show that models have varied performance over different metrics, like GPT-4V-Turbo is more robust to correctness hallucinations and Gemini is more robust to consistent hallucinations.
Our results affirm the effectiveness of our pipeline in crafting hallucination cases with a high attack success rate, while using the same model for object prompting and VQA tasks usually causes more hallucinations due to inherited biases. We attribute this phenomenon to the diversity of the prior across different LVLMs as the VQA model may find the object prompted by other LVLMs less abnormal and it is less likely to suffer from hallucinations by this prompted object.


\paragraph{Object-scene Alignment.}
Table~\ref{ablation_random} presents results using different object retrieval policies under object insertion experiments using GPT-4V-Turbo, including abnormal (intentionally chooses irrelevant objects), random (randomly chooses objects), and same (chooses objects aligned with the existing contexts in the image). Results show that the abnormal object insertion strategy shows a significantly high ASR over questions probing the existence of perturbed objects, and the same object insertion strategy shows a greatly lower overall MASR. As the object retrieval and insertion strategy mainly affects the LVLMs' ability to identify the perturbed objects from the image, abnormal object insertion more easily causes the cognitive disorder of LVLMs, reflected by the high MASR values. On the other hand, LVLMs are more likely to make correct predictions when the perturbed objects are contextually aligned with the image, leading to a lower MASR value.

\begin{table}[t]
  \begin{center}
  \resizebox{\linewidth}{!}{
    \begin{tabular}{lccccccccccccccccc}
    \toprule
        \multicolumn{1}{c}{} & \multicolumn{3}{c}{\makecell{Overall}}
        & \multicolumn{3}{c}{\makecell{Existence}} & \multicolumn{3}{c}{\makecell{Spatial Relation}} \\
        \cmidrule(lr){2-4} \cmidrule(lr){5-7} \cmidrule(lr){8-10}
         Alignment & \makecell{Overall \\ ASR} & \makecell{Overall \\ MASR} & \makecell{Overall \\ CASR} & \makecell{Exi. \\ ASR} & \makecell{Exi. \\ MASR} & \makecell{Exi. \\ CASR} & \makecell{Sp. \\ ASR} & \makecell{Sp. \\ MASR} & \makecell{Sp. \\ CASR}\\
        \midrule
        Abnormal & 96.0 & 80.0 & 92.5 & 93.0 & 62.0 & 88.5 & 78.1 & 71.2 & 60.6\\
        Random & 98.5 & 82.0 & 93.5 & 91.5 & 50.5 & 89.0 & 84.0 & 74.9 & 59.4\\
        Same & 93.0 & 65.5 & 90.0 & 88.0 & 27.5 & 85.5 & 83.1 & 70.9 & 62.2 \\
    \hline
    \end{tabular}
    }
    \end{center}
    \vspace{-3mm}
     \caption{\textbf{Ablation on object-scene alignments with abnormal object insertion using GPT-4V-Turbo.}} 
    \label{ablation_random}
    \vspace{-3mm}
\end{table}

\begin{table*}[h]
  \begin{center}
  \resizebox{\textwidth}{!}{
    \begin{tabular}{ccccccccccccc}
    \toprule
        \multicolumn{2}{c}{} & \multicolumn{2}{c}{\makecell{Diversity (200 / 126 samples)}}  & \multicolumn{3}{c}{\makecell{Image Quality}}
        & \multicolumn{3}{c}{\makecell{Effectiveness of Exi. Questions}} & \multicolumn{3}{c}{\makecell{Effectiveness of Sp. Questions}}\\
        \cmidrule(lr){3-4} \cmidrule(lr){5-7} \cmidrule(lr){8-13}
        \makecell{Strategy} & \makecell{Data} &  \makecell{\# Scene $\uparrow$} & \makecell{\# Obj. $\uparrow$} & \makecell{Origin \\ IS $\uparrow$} & \makecell{Edited \\ IS $\uparrow$} & \makecell{FID $\downarrow$} & \makecell{Overall \\ Avg. \# Q. $\downarrow$} & \makecell{Avg. \# Q. \\ Correctness $\downarrow$} & \makecell{Avg. \# Q. \\ Consistency $\downarrow$} & \makecell{Overall \\ Avg. \# Q. $\downarrow$} & \makecell{Avg. \# Q. \\ Correctness $\downarrow$} & \makecell{Avg. \# Q. \\ Consistency $\downarrow$}\\
        \midrule
        \multirow{3}*{\makecell{Abnormal Obj. \\ Insertion}} & Synthetic & 152 & 89 & \makecell{4.977 \\ $\pm$ 0.754} & \makecell{5.295 \\ $\pm$ 0.988} & 161.56 & 2.162 & 3.243 & 1.622 & 2.318 & 2.134 & 2.537\\
        ~ & Real-World & 118 & 78 & \makecell{5.126 \\ $\pm$ 0.715} & \makecell{5.143 \\ $\pm$ 0.865} & 162.33 & 1.780 & 2.268 & 1.465 & 1.855 & 1.801 & 1.913\\
        \midrule
        \multirow{3}*{\makecell{Paired Obj. \\ Insertion}} & Synthetic & 165 & 70 &  \makecell{5.936 \\ $\pm$ 1.230} & \makecell{6.211 \\ $\pm$ 1.146} & 152.99 & 2.444 & 3.822 & 1.796 & 2.822 & 2.511 & 3.220\\
        ~ & Real-World & 118 & 78 & \makecell{5.741 \\ $\pm$ 0.723} & \makecell{5.965 \\ $\pm$ 0.754} & 119.73 & 2.114 & 3.321 & 1.550 & 2.003 & 1.820 & 2.227\\
        \midrule
        \multirow{3}*{\makecell{Correlated Obj. \\ Removal}} & Synthetic & 193 & N/A & \makecell{5.455 \\ $\pm$ 0.834} & \makecell{5.529 \\ $\pm$ 0.895} & 220.15 & 3.241 & 2.388 & 4.255 & 1.717 & 1.968 & 1.523\\
        ~ & Real-World & 118 & 78 & \makecell{5.924 \\ $\pm$ 0.575} & \makecell{5.664 \\ $\pm$ 0.957} & 363.93 & 2.927 & 2.369 & 3.472 & 1.725 & 1.898 & 1.581\\
    \hline
    \end{tabular}
    }
    \end{center}

    \caption{\textbf{Metrics to evaluate the diversity, quality, and effectiveness of the benchmark generated by our \ours{}.} We assess the benchmark based on the following three aspects: \textbf{(1) Diversity} indicates the number of different scenes and objects to construct the dataset, out of 200 samples; \textbf{(2) Image Quality} is represented using the Inception Score~\cite{salimans2016improved} for both original and edited images, and Frechet Inception Distance~\cite{heusel2017gans} between the original and edited images; \textbf{(3) Effectiveness} is measured by the average number of questions constructed to induce hallucination.}

    
    \label{benchmark_metrics}
    \vspace{-3mm}
\end{table*}

\begin{table}[h]
  \begin{center}
  \resizebox{\columnwidth}{!}{
    \begin{tabular}{lccccccc}
    \toprule
        ~ & ~ & \multicolumn{3}{c}{\makecell{Synthetic Data}}
        & \multicolumn{3}{c}{\makecell{Real-World Data}} \\
        \cmidrule(lr){3-5} \cmidrule(lr){6-8} 
        \makecell{LVLMs} &  \makecell{Overall \\ Acc.} & \makecell{Overall \\ Acc.} & \makecell{Exi. \\ Acc.} & \makecell{Sp. \\ Acc.} & \makecell{Overall \\ Acc.} & \makecell{Exi. \\ Acc.} & \makecell{Sp. \\ Acc.}\\
        \midrule
        GPT-4V-Turbo \cite{yang2023dawn} & \textbf{66.0} & \textbf{68.5} & \textbf{68.3} & \textbf{68.8} & \textbf{62.9} & \textbf{71.5} & \textbf{56.3} \\
        Gemini Pro Vision \cite{google_gemini} & 51.4 & 53.5 & 59.4 & 43.4 & 48.8 & 70.6 & 31.8\\
        Claude 3 \cite{claude3} & 37.1 & 37.2 & 44.6 & 24.7 & 36.9 & 55.6 & 22.4\\
        LLaVA-1.5 \cite{liu2023improved} & 44.5 & 46.6 & 54.2 & 33.8 & 41.8 & 60.4 & 27.3 \\
        miniGPT4 \cite{zhu2023minigpt} & 51.0 & 50.2 & 56.4 & 39.7 & 52.1 & 67.7 & 39.9\\
    \hline
    \end{tabular}
    }
    \end{center}
    \caption{\textbf{Leaderboard of SOTA LVLMs on the benchmark created by \ours{}.} We highlight the top-performing model.
    }
    \label{leaderboard}
    \vspace{-3mm}
\end{table}

\subsection{Benchmark Curation}
\label{Sec: Benchmark Curation}

Based on the data generated from our experiments and following a thorough manual inspection, we developed a benchmark for further evaluation. We generate the benchmark over all three hallucinations crafting strategies we proposed, using both scene and object images from synthetic and real-world datasets. 
Table~\ref{benchmark_metrics} presents the evaluation results for all metrics. For diversity, we sampled 200 hallucination examples from the synthetic dataset and 126 from the real-world dataset. The image quality results indicate that the crafted images have quality comparable to both datasets, with FID scores showing close distribution for abnormal object insertion and paired object insertion, while object removal may cause significant variance. The effectiveness metric indicates an average of 2.5 questions to induce hallucination.

We provide a leaderboard of SOTA LVLMs on the benchmark created by \ours{} in Table~\ref{leaderboard}, while LVLMs are evaluated on both synthetic and real-world data. Metrics include the question-answering accuracy of each LVLM, covering synthetic and real-world data as well as the existence and spatial relation questions. The evaluation results show: (1) GPT-4V-Turbo has the highest accuracy among all LVLMs evaluated; (2) All LVLMs perform better on benchmarks made by synthetic data than real-world data; (3) LVLMs are more robust to hallucinations induced by existence questions than spatial relation questions.

\section{Conclusion}

In this paper, we introduce \ours{}, the first automatic benchmark generation approach to create diverse hallucination examples. Inspired by \textit{schema} in cognitive science, we analyze the mechanism of how and when LVLM hallucinations are triggered. We then reverse-engineer the hallucinating images based on probed LVLMs' language priors by three principal strategies, abnormal object insertion, paired object insertion, and correlated object removal, that manipulate scene images using object insertion or removal to create conflicts with the priors. 
We construct textual probing methods to construct and detect hallucinations created. \ours{} achieves a significant success rate of inducing LVLM hallucinations on manipulating both synthetic and real-world data. 
We will keep improving the quality of the synthesized images by inpainting techniques based on more recent text-to-image models. 
Meanwhile, we will explore better textual probing methods extracting more diverse contextual information within the image. 
We will also further investigate the causes of multi-modal hallucinations and build a more rigorous mathematical model for them.

\newpage
\section{Limitation}
A limitation of our current image manipulation strategies lies on the object insertion, where we are using a primitive image stitch pipeline to insert prompted objects into the scene image. Though the success of this strategy is supported by the experimental results, the edited images have strong perceivable hand-crafting evidences which lower the quality of the resulted hallucinating images. Another limitation comes from the diversity of questions, as they mainly focus on objects' existence and spatial relations but have not explore the objects' attributes, e.g., color, pattern, and conditions, on which hallucinations might also emerge. We will take efforts to overcome them in our future update of \ours{}.

\section*{Acknowledgments}
We would like to thank Qingyuan Chen (New York University) for helpful discussions and feedback about hallucination cognition science in this paper.
This material is based upon work supported by the National Science
Foundation under Grant No. {iis}-2403436 (Boyd-Graber).
Any opinions, findings, and conclusions or recommendations expressed
in this material are those of the author(s) and do not necessarily
reflect the views of the National Science Foundation.

\bibliography{reference}

\clearpage

\appendix

\section{More Experimentation Results}
\label{appendix:exp}

\begin{table*}[h]
  \begin{center}
  \resizebox{\textwidth}{!}{
    \begin{tabular}{clccccccccccccc}
    \toprule
        \multicolumn{2}{c}{} & \multicolumn{3}{c}{\makecell{Overall}}
        & \multicolumn{3}{c}{\makecell{Existence}}  & \multicolumn{3}{c}{\makecell{Spatial Relation}} \\
        \cmidrule(lr){3-5} \cmidrule(lr){6-8} \cmidrule(lr){9-11} 
         \makecell{Manipulation \\ Strategies} & LVLMs &  \makecell{Overall \\ ASR} & \makecell{Overall \\ MASR} & \makecell{Overall \\ CASR} & \makecell{Exi. \\ ASR} & \makecell{Exi. \\ MASR} & \makecell{Exi. \\ CASR} & \makecell{Sp. \\ ASR} & \makecell{Sp. \\ MASR} & \makecell{Sp. \\ CASR}\\
        \midrule
        \multirow{5}*{\makecell{Abnormal Obj. \\ Insertion}} & GPT-4V-Turbo \cite{yang2023dawn} & 96.0 & 80.0 & 92.5 & 93.0 & 62.0 & 88.5 & 78.1 & 71.2 & 60.6  \\
        ~ & Gemini Pro Vision \cite{google_gemini} & 97.0 & 90.5 & 90.0 & 84.5 & 75.5 & 68.0 & 89.1 & 81.0 & 73.6 \\
        ~ & Claude 3 \cite{claude3} & 97.4 & 90.7 & 96.0 & 95.3 & 81.5 & 90.7 & 92.3 & 79.2 & 90.8  \\
        ~ & LLaVA-1.5 \cite{liu2023improved} & 97.7 & 94.2 & 94.0 & 97.9 & 87.4 & 95.6 & 96.2 & 83.3 & 97.6 \\
        ~ & miniGPT4 \cite{zhu2023minigpt} & 98.1 & 95.1 & 98.0 & 98.1 & 89.8 & 97.7 & 97.1 & 89.3 & 98.2\\
        \midrule
        \multirow{5}*{\makecell{Paired Obj. \\ Insertion}} & GPT-4V-Turbo \cite{yang2023dawn} & 99.5 & 93.5 & 97.0 & 91.5 & 60.5 & 86.0 & 81.7 & 72.0 & 58.3 \\
        ~ & Gemini Pro Vision \cite{google_gemini} & 100.0 & 100.0 & 99.5 & 99.5 & 99.5 & 97.5 & 85.7 & 62.3 & 74.0\\
        ~ & Claude 3 \cite{claude3} & 100.0 & 99.0 & 99.0 & 99.0 & 86.0 & 98.0 &  95.5 & 91.0 & 91.0 \\
        ~ & LLaVA-1.5 \cite{liu2023improved} &  99.7 & 95.1 & 98.9 & 97.6 & 98.4 & 94.1 & 81.8 & 79.7 & 72.3\\
        ~ & miniGPT4 \cite{zhu2023minigpt} & 100.0 & 99.8 & 100.0 & 99.1 & 99.3 & 99.7 & 83.9 & 71.1 & 75.2\\
        \midrule
        \multirow{5}*{\makecell{Correlated Obj. \\ Removal}} & GPT-4V-Turbo \cite{yang2023dawn} & 93.0 & 84.0 & 84.0 & 69.5 & 68.5 & 46.0 &  85.5 & 67.6 & 79.2 \\
        ~ & Gemini Pro Vision \cite{google_gemini} & 95.0 & 92.0 & 93.0 & 77.0 & 77.0 & 70.5 &  91.1 & 83.2 & 87.4 \\
        ~ & Claude 3 \cite{claude3} & 99.0 & 98.0 & 89.0 & 92.0 & 92.0 & 64.0 &  88.5 & 83.3 & 82.3 \\
        ~ & LLaVA-1.5 \cite{liu2023improved} & 97.1 & 88.9 & 87.4 & 70.8 & 71.4 & 65.3 &  87.4 & 75.3 & 86.9 \\
        ~ & miniGPT4 \cite{zhu2023minigpt} & 96.7 & 90.1 & 91.5 & 72.9 & 72.7 & 63.7 &  86.7 & 76.4 & 85.5\\
    \hline
    \end{tabular}
    }
    \end{center}
    \caption{\textbf{Attack Results across all LVLMs with three manipulation strategies on synthetic data.}} 
    \label{basic_results}
\end{table*}

\begin{table*}[h]
  \begin{center}
  \resizebox{\textwidth}{!}{
    \begin{tabular}{clccccccccccccc}
    \toprule
        \multicolumn{2}{c}{} & \multicolumn{3}{c}{\makecell{Overall}}
        & \multicolumn{3}{c}{\makecell{Existence}} & \multicolumn{3}{c}{\makecell{Spatial Relation}} \\
        \cmidrule(lr){3-5} \cmidrule(lr){6-8} \cmidrule(lr){9-11} 
        \makecell{Manipulation \\ Strategies} & LVLMs &  \makecell{Overall \\ ASR} & \makecell{Overall \\ MASR} & \makecell{Overall \\ CASR} & \makecell{Exi. \\ ASR} & \makecell{Exi. \\ MASR} & \makecell{Exi. \\ CASR} & \makecell{Sp. \\ ASR} & \makecell{Sp. \\ MASR} & \makecell{Sp. \\ CASR}\\
        \midrule
        \multirow{5}*{\makecell{Abnormal Obj. \\ Insertion}} & GPT-4V-Turbo \cite{yang2023dawn} & 96.0 & 80.0 & 92.5 & 93.0 & 62.0 & 88.5  & 78.1 & 71.2 & 60.6 \\
        ~ & Gemini Pro Vision \cite{google_gemini} & 89.5 & 82.5 & 76.5 & 80.5 & 66.5 & 64.5  & 78.8 & 66.3 & 60.6 \\
        ~ & Claude 3 \cite{claude3} & 97.0 & 93.0 & 95.0 & 94.0 & 82.0 & 90.0  & 90.1 & 84.6 & 86.8 \\
        ~ & LLaVA-1.5 \cite{liu2023improved} & 96.1 & 79.4 & 83.3 & 91.7 & 70.5 & 81.4 &  72.2 & 68.1 & 60.4\\
        ~ & miniGPT4 \cite{zhu2023minigpt} & 95.5 & 72.1 & 70.9 & 82.7 & 61.8 & 77.2  & 74.1 & 70.5 & 65.8\\
        \midrule
        \multirow{5}*{\makecell{Paired Obj. \\ Insertion}} & GPT-4V-Turbo \cite{yang2023dawn} & 99.5 & 93.5 & 97.0 & 91.5 & 60.5 & 86.0 & 81.7 & 72.0 & 58.3\\
        ~ & Gemini Pro Vision \cite{google_gemini} & 100.0 & 90.5 & 99.0 & 83.5 & 67.0 & 67.0 & 78.3 & 58.3 & 56.0\\
        ~ & Claude 3 \cite{claude3} & 100.0 & 97.0 & 100.0 & 99.0 & 89.0 & 99.0 & 94.2 & 86.0 & 90.7 \\
        ~ & LLaVA-1.5 \cite{liu2023improved} & 100.0 & 96.1 & 98.7 & 90.3 & 64.1 & 87.0 & 84.4 & 70.2 & 57.9\\
        ~ & miniGPT4 \cite{zhu2023minigpt} & 100.0 & 97.7 & 99.6 & 92.7 & 78.2 & 89.7 & 87.8 & 80.1 & 67.5\\
        \midrule
        \multirow{5}*{\makecell{Correlated Obj. \\ Removal}} & GPT-4V-Turbo \cite{yang2023dawn} & 93.0 & 84.0 & 84.0 & 69.5 & 68.5 & 46.0  & 85.5 & 67.6 & 79.2\\
        ~ & Gemini Pro Vision \cite{google_gemini} & 97.0 & 94.0 & 90.5 & 74.5 & 74.5 & 60.5  & 91.9 & 83.2 & 89.0\\
        ~ & Claude 3 \cite{claude3} & 100.0 & 100.0 & 93.0 & 94.0 & 94.0 & 66.0  & 90.4 & 84.3 & 89.2\\
        ~ & LLaVA-1.5 \cite{liu2023improved} & 98.1 & 91.2 & 89.8 & 70.9 & 69.9 & 54.1  & 87.2 & 76.1 & 78.8\\
        ~ & miniGPT4 \cite{zhu2023minigpt} & 97.9 & 93.5 & 91.6 & 78.3 & 68.1 & 57.9  & 89.3 & 77.4 & 82.1\\
    \hline
    \end{tabular}
    }
    \end{center}
    \caption{\textbf{Attack Results across all LVLMs with three manipulation strategies using the same synthetic dataset.} This aligned synthetic dataset was created by GPT-4V-Turbo and DALL-E-3, and is used for all victim LVLMs.} 
    \label{reuse_results}
\end{table*}

\begin{table*}[h]
  \begin{center}
  \resizebox{\textwidth}{!}{
    \begin{tabular}{clccccccccccccc}
    \toprule
        \multicolumn{2}{c}{} & \multicolumn{3}{c}{\makecell{Overall}}
        & \multicolumn{3}{c}{\makecell{Existence}} & \multicolumn{3}{c}{\makecell{Spatial Relation}} \\
        \cmidrule(lr){3-5} \cmidrule(lr){6-8} \cmidrule(lr){9-11}
        \makecell{Manipulation \\ Strategies} & LVLMs &  \makecell{Overall \\ ASR} & \makecell{Overall \\ MASR} & \makecell{Overall \\ CASR} & \makecell{Exi. \\ ASR} & \makecell{Exi. \\ MASR} & \makecell{Exi. \\ CASR} & \makecell{Sp. \\ ASR} & \makecell{Sp. \\ MASR} & \makecell{Sp. \\ CASR}\\
        \midrule
        \multirow{5}*{\makecell{Abnormal Obj. \\ Insertion}} & GPT-4V-Turbo \cite{yang2023dawn} & 100.0 & 98.4 & 98.4 & 97.6 & 74.2 & 92.7 &   97.5 & 92.7 & 89.5\\
        ~ & Gemini Pro Vision \cite{google_gemini} & 100.0 & 100.0 & 97.6 & 97.6 & 94.3 & 89.4 &  94.3 & 86.2 & 78.9\\
        ~ & Claude 3 \cite{claude3} &  100.0 & 100.0 & 100.0 & 100.0 & 91.3 & 100.0 &  98.4 & 96.8 & 98.4\\
        ~ & LLaVA-1.5 \cite{liu2023improved} & 100.0 & 100.0 & 98.9 & 98.6 & 89.2 & 92.5 &  95.9 & 91.7 & 92.6\\
        ~ & miniGPT4 \cite{zhu2023minigpt} &  100.0 & 100.0 & 97.9 & 98.0 & 90.5 & 92.6 &  96.1 & 93.1 & 87.5\\
        \midrule
        \multirow{5}*{\makecell{Paired Obj. \\ Insertion}} & GPT-4V-Turbo \cite{yang2023dawn} & 100.0 & 100.0 & 99.2 & 99.2 &64.5 & 95.2 & 100.0 & 97.6 & 87.9\\
        ~ & Gemini Pro Vision \cite{google_gemini} & 100.0 & 99.2 & 100.0 & 99.2 & 84.0 & 89.6 & 90.4 & 84.0 & 68.8\\
        ~ & Claude 3 \cite{claude3} & 100.0 & 99.2 & 100.0 & 97.6 & 64.3 & 96.8 & 99.2 & 98.4 & 99.2\\
        ~ & LLaVA-1.5 \cite{liu2023improved} & 99.7 & 98.5 & 99.3 & 94.5 & 61.8 & 89.0 & 97.8 & 95.6 & 84.9\\
        ~ & miniGPT4 \cite{zhu2023minigpt} & 100.0 & 100.0 & 99.5 & 99.5 & 65.7 & 96.1 & 99.8 & 98.6 & 89.1\\
        \midrule
        \multirow{5}*{\makecell{Correlated Obj. \\ Removal}} & GPT-4V-Turbo \cite{yang2023dawn} & 94.4 & 88.0 & 84.0 & 75.2 & 72.8 & 55.2 &  85.4 & 73.4 & 81.5\\
        ~ & Gemini Pro Vision \cite{google_gemini} & 96.8 & 95.2 & 92.0 & 77.6 & 77.6 & 68.0 &  94.2 & 86.0 & 89.3\\
        ~ & Claude 3 \cite{claude3} & 98.4 & 98.4 & 94.4 & 96.0 & 95.2 & 74.6 &  89.6 & 88.0 & 88.8\\
        ~ & LLaVA-1.5 \cite{liu2023improved} & 93.1 & 97.6 & 94.6 & 78.1 & 73.7 & 71.1 &  95.7 & 89.3 & 90.1\\
        ~ & miniGPT4 \cite{zhu2023minigpt} & 97.8 & 96.3 & 89.1 & 76.9 & 73.4 & 70.4 &  87.8 & 74.9 & 87.5\\
    \hline
    \end{tabular}
    }
    \end{center}

      \caption{\textbf{Attack Results across all LVLMs with three manipulation strategies on a real-world dataset.} The real-world data is created from the Common Objects in Context (COCO) dataset validation set~\cite{lin2014microsoft}.} 
    \label{real_results}
\end{table*}
\begin{table*}[h]
    \centering
    \resizebox{0.8\textwidth}{!}{
    \renewcommand\arraystretch{1.5}
    \begin{tabular}{l|l|p{26em}}
    \hline
    \textbf{Category} & \textbf{Contextual Info.} & \textbf{Question}\\
    \hline
    \multirow{2}*{\makecell{Existence}} & N/A & \textit{Is there a $\{Target Object Name \}$ in this image?} \\
    ~ & Image-level Caption & \textit{We have an image depicting $\{ Image Caption \}$. Is there a $\{Target Object Name \}$ in this image?} \\
    \hline
    \multirow{4}*{\makecell{Correlation}} & N/A & \textit{Is there a $\{ Object Name \}$ in this image?} \\
    ~ & Paired Obj. & \textit{We have $\{Paired Object Name \}$ in this image. Is there a $\{ Object Name \}$ in this image?} \\
    ~ & Image-level Caption & \textit{We have an image depicting $\{ Image Caption \}$. Is there a $\{ Object Name \}$ in this image?} \\
    \hline
    \multirow{4}*{\makecell{Spatial \\ Relation}} & N/A & \textit{Is the $\{Target Object Name \}$ $\{spatial relation \}$ a/an $\{Existing Object Name \}$ in this image, given their center positions?} \\
    ~ & Obj. Description & \textit{Is the object ($\{Target Object Description \}$) $\{spatial relation \}$ a/an $\{Existing Object Name \}$ in this image, given their center positions?} \\
\hline
    \end{tabular}
    }
    \caption{Questions Constructed to Induce Hallucinations}
    \label{app:question_detail}
\end{table*}

\subsection{Synthetic Dataset}
\label{app: Synthetic Dataset}
Table~\ref{basic_results} presents the results of victim LVLMs under three attack strategies using synthetic datasets.
GPT-4V-Turbo exhibits the highest robustness to hallucination attacks among all strategies, particularly showing stronger resistance to correctness hallucinations than to consistency hallucinations. Open-source, smaller LVLMs like LLaVA-1.5 and miniGPT4 perform comparably to Gemini and better than Claude.
Questions probing the existence of objects are easier to cause hallucination of LVLMs than those probing spatial relations. Question targeting to inserted objects, including the existence questions of abnormal and paired object insertions, contributes to a better hallucination attack success rate than those targeting hypothetical objects, like correlation questions and existence questions for correlated object removal.
Hallucination attacks exploiting inconsistencies in responses are more effective for existence questions about inserted objects and spatial relation queries but are less effective for questions about object removal.
Results demonstrate that using sequences of questions to probe hallucinations with varying contextual information from the image effectively disrupts the cognitive processing of LVLMs, showing superior results compared to strategies that involve object removal to induce expectation violations in LVLMs.



\subsection{Aligned Synthetic Dataset}
\label{app: Aligned Synthetic Dataset}
In an ablation study, we assessed the vulnerability of various LVLMs to three attack strategies using the same synthetic datasets, which incorporate abnormal objects and scene images generated by GPT-4V-Turbo and DALL-E-3. The results, detailed in Table~\ref{reuse_results}, indicate that all LVLMs, except Claude, show a decrease in both MASR and CASR for existence and spatial relation questions, but an increase in attack success rate for correlation questions. This suggests that LVLMs exhibit stronger resistance to hallucinations induced by images from other models than by those they generated themselves, corroborating findings in Section~\ref{Sec: Ablation Studies}. GPT-4V-Turbo, in particular, excels in handling paired object insertions. We attribute these differences to the varying priors among LVLMs; a VQA model may perceive an object suggested by another LVLM as less abnormal or correlated, thus reducing the likelihood of hallucinations. Further insights are explored in our ablation study in Section~\ref{Sec: Ablation Studies}, where we swap the roles of LVLMs in object prompting and VQA tasks to examine the impact of using different LVLMs for these functions.



\subsection{Real-world Dataset}
\label{app: Real-world Dataset}
Table~\ref{real_results} displays the results of hallucination attacks using real-world datasets across three strategies. The results indicate that victim LVLMs are more susceptible to hallucination attacks with real-world datasets, showing increased success rates for all metrics compared to those in Table~\ref{basic_results}. We hypothesize the discrepancy in LVLMs' performance over synthetic and real-world datasets comes from their lack of ability to address the complexity and diversity within the real-world data, which causes its higher vulnerability to our attack strategies when using real-world data.



\section{Discussion}
\label{appendix: discussion}



Across all results discussed in Sections~\ref{Sec: Main Results}, \ref{Sec: Ablation Studies}, and Appendix~\ref{appendix:exp}, we identified key insights into our proposed strategies and LVLMs' resistance to hallucination attacks.

\textbf{Robust to Absence, Vulnerable to Perturbation.} LVLMs are more vulnerable to hallucinations involving object insertions, such as abnormal and paired object insertion strategies, compared to those focused on object absence, like in the correlation object removal strategy. This suggests that attacks leveraging cognitive dissonance through object insertion are more effective than those relying on expectancy violations via object removal.

\textbf{Robustness to Hallucination Attacks across LVLMs.} GPT-4V shows superior resistance to the hallucination attacks we proposed, especially in the MASR metric assessing correctness hallucinations. Gemini slightly outperforms other LVLMs in the CASR metric. Larger models like GPT-4V-Turbo, Gemini Pro Vision, and Claude 3 generally surpass smaller ones such as LLaVA-1.5 and miniGPT4, demonstrating a link between model size and hallucination resistance.



\textbf{Real-world Data Increases Difficulty.} Victim LVLMs show increased vulnerability to hallucination attacks with real-world datasets than synthetic ones. Real-world images generally contain more contextual information than synthetic ones, causing LVLMs to struggle with the added complexity and diversity, thus heightening their vulnerability to hallucination attacks based on real-world data.

\textbf{Swap Object Prompting and VQA Model Help.} According to results in Fig.~\ref{fig:heatmap} and Appendix~\ref{app: Aligned Synthetic Dataset}, utilizing different LVLMs to prompt objects for image manipulation and handle VQA tasks reduces hallucinations. This effect is attributed to the varying priors among LVLMs; different models may have different responses to prompted objects for insertion or removal, making some LVLMs more resistant to hallucination cases generated by another model.
\section{GPT4-Assisted Evaluation}
\label{Appendix: eval prompt}

We evaluated the accuracy of the LVLMs' output by using the following prompt in GPT-4:\looseness-1

\textit{"Imagine you are an intelligent teacher. Thoroughly read the question, reference answer and the prediction answer to ensure a clear understanding of the information provided. Assess the correctness of the predictions. If the prediction answer does not conflict with the reference answer, please generate “correct”. If the prediction answer conflict with the reference answer, please generate “incorrect”. The output should only be “correct” or “incorrect”. The question, model generated response and ground truth is as follows: \textellipsis"}
\section{Question Details}


Table~\ref{app:question_detail} outlines the details of the questions constructed to probe hallucinations. As outlined in Section~\ref{Methodology: Question Construction} and~\ref{Methodology: Hallucination Detection}, we employ a series of questions varying in contextual information to explore hallucinations. For questions probing the existence of the target object, we create queries both with and without image-level captions. For those probing the correlation of paired objects, we provide three levels of contextual information: none, the existence of the paired object, and image-level captions. For spatial relation probes, questions utilize the target object’s name and descriptive text.

Under each category, we examine conflicts among questions with varying contexts to detect potential consistency in hallucinations.

\newpage
\section{More Examples}
\label{Appendix: Showcase}
We provide several showcases across all $3$ hallucination crafting strategies and all questions covered by \ours{}. \textbf{Each figure is self-contained for readability}, where we highlight the control pairs, the responses of GPT-4V and LLaVA-1.6, the failures of those models, and the corresponding part of the answers.

Fig.~\ref{Abnormal Object Insertion -- Existence} and~\ref{Abnormal Object Insertion -- Spatial Relation} display cases from the abnormal object insertion strategy. Fig.~\ref{Abnormal Object Insertion -- Existence} illustrates both GPT-4V and LLaVA-1.6 inconsistently answering the existence of an inserted object. In Fig.\ref{Abnormal Object Insertion -- Spatial Relation}, only GPT-4V experiences correctness hallucination, while LLaVA-1.6 responds accurately.


Fig.~\ref{Paired Object Insertion -- Correlated 1} and \ref{Paired Object Insertion -- Correlated 2} exhibit cases from the paired object insertion strategy, focusing on the absence of one object paired with an existing object. Fig.~\ref{Paired Object Insertion -- Correlated 1} shows GPT-4V failing to provide consistent answers across varying contexts, whereas LLaVA-1.6 answers correctly and consistently. In Fig.~\ref{Paired Object Insertion -- Correlated 2}, both models show correctness hallucinations and inconsistency in responses concerning the existence of the paired object.



For hallucination cases made by correlated object removal, Fig.~\ref{Correlated Object Removal -- Existence} shows that both models fail to make correct answers to all questions, while GPT-4V makes wrong answers to both questions and LLaVA-1.6 makes inconsistent answers over questions. The example in Fig.~\ref{Correlated Object Removal -- Spatial Relation} shows that both LVLMs fail to make consistent answers to the spatial relation between the removed object and one of the existing objects in the edited image as they mistakenly assume the existence of the removed object given the contexts presented in the image.

\section{Failure Case}
\label{Appendix: failure cases}
\begin{figure*}
\begin{center}
\begin{tikzpicture}
\node [mybox,title=Failure Case: Non-Perceivable Objects Prompted] (box){%
    \begin{minipage}{2\columnwidth}
        \raggedright
        \bluebold{Scene Name: } \textit{Hiking, horseback, wilderness} \\
        \bluebold{Detected Objects: } \textit{Man riding horse}, \textit{Large gray rock}, \textit{Riding horse forward}, \textit{Red bandana, glasses} \\
        \bluebold{Inserted Object: } \textit{Refrigerator} \\
        \bluebold{Image Manipulation: } \textit{Refrigerator} is inserted into the image.
        \\[2pt]

    \begin{multicols}{2}

        \centering
        \includegraphics[width=0.6\columnwidth]{failure_cases/obj/1/init.png}      \\
        \textit{\textbf{Original image}} \\        
        \columnbreak
        \centering
        \includegraphics[width=0.6\columnwidth]{failure_cases/obj/1/results.png} \\
        \textit{\textbf{Edited image}} \\
    \end{multicols}

        \raggedright
        \bluebold{Attack: } Adding new irrelevant object \textit{Refrigerator} -- Existence \\
        \bluebold{Question: } Is there a/an refrigerator in this image? \\
        \bluebold{GPT-4V: } No, there is no refrigerator in this image. The image shows a man wearing a bandana looking at another man riding a horse, which is carrying what appear to be large wooden boxes or panels. They are on a rocky trail in a forested area. \\
        \bluebold{LLaVA-1.6: } No, there is no refrigerator in this image. The image shows a person riding a horse, with another person walking alongside, and a third person looking back at the camera. They are in a natural setting with trees and rocks. \\
        \quad \\ 
    \par\noindent\rule[15pt]{\textwidth}{1pt}\\[-30pt]
    \quad \\ 
    \raggedright
        \bluebold{Scene Name: } \textit{Fishing boats, marshland} \\
        \bluebold{Detected Objects: } \textit{Harbor sailing boats}, \textit{Marshland with egrets}, \textit{Blurred swan figure}, \textit{Fishing boat masts} \\
        \bluebold{Inserted Object: } \textit{Airplane} \\
        \bluebold{Image Manipulation: } \textit{Airplane} is inserted into the image.
        \\[2pt]

    \begin{multicols}{2}

        \centering
        \includegraphics[width=0.6\columnwidth]{failure_cases/obj/2/init.png}      \\
        \textit{\textbf{Original image}} \\        
        \columnbreak
        \centering
        \includegraphics[width=0.6\columnwidth]{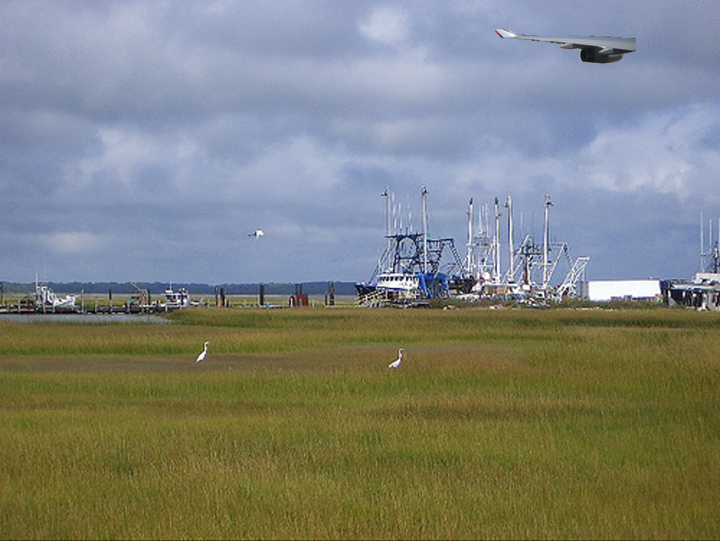} \\
        \textit{\textbf{Edited image}} \\
    \end{multicols}

        \raggedright
        \bluebold{Attack: } Adding new irrelevant object \textit{Airplane} -- Existence \\
        \bluebold{Question: } Is there a/an airplane in this image? \\
        \bluebold{GPT-4V: } Yes, there is an airplane in the image. It is flying in the upper left part of the photo, above the harbor area and the grassy field.\\
        \bluebold{LLaVA-1.6: } Yes, there is an airplane in the image. It appears to be flying in the sky above the landscape.\\
    
    \end{minipage}
};
\end{tikzpicture}%
\caption{\textbf{Failure Case: Non-Perceivable Objects Prompted.}}
\label{Failure Case: Non-Perceivable Objects Prompted}
\end{center}
\end{figure*}

\begin{figure*}
\begin{center}
\begin{tikzpicture}
\node [mybox,title=Failure Case: When LVLMs Detect the Edition] (box){%
    \begin{minipage}{2\columnwidth}
        \raggedright
        \bluebold{Scene Name: } \textit{Bright, tidy kitchen} \\
        \bluebold{Detected Objects: } \textit{White kitchen dishwasher}, \textit{White vertical board}, \textit{Wooden kitchen cabinets}, \textit{White kitchen dishwasher}, \textit{Bowl of oranges} \\
        \bluebold{Inserted Object: } \textit{Giraffe} \\
        \bluebold{Image Manipulation: } \textit{Giraffe} is inserted into the image.
        \\[2pt]

    \begin{multicols}{2}

        \centering
        \includegraphics[width=0.6\columnwidth]{failure_cases/edit/1/init.png}      \\
        \textit{\textbf{Original image}} \\        
        \columnbreak
        \centering
        \includegraphics[width=0.6\columnwidth]{failure_cases/edit/1/results.png} \\
        \textit{\textbf{Edited image}} \\
    \end{multicols}

        \raggedright
        \bluebold{Attack: } Adding new irrelevant object \textit{Giraffe} -- Existence \\
        \bluebold{Question: } Is there a/an giraffe in this image? \\
        \bluebold{GPT-4V: } There isn't an actual giraffe in the image. The image includes a graphical or edited-in representation of a giraffe in a kitchen setting, which is a fictional and humorous element because giraffes are not indoor animals nor they are typically found in such environments. \\
        \bluebold{LLaVA-1.6: } Yes, there is a giraffe in this image. It appears to be a digitally added or edited element, as it is not possible for a giraffe to be in a kitchen setting like this. \\
        \quad \\ 
    \par\noindent\rule[15pt]{\textwidth}{1pt}\\[-30pt]
    \quad \\ 
    \raggedright
        \bluebold{Scene Name: } \textit{Mountain cottage} \\
        \bluebold{Detected Objects: } \textit{Burning wooden logs}, \textit{Snowshoes hanging outside}, \textit{Snowy pine tree}, \textit{Enclosed wood fireplace} \\
        \bluebold{Inserted Object: } \textit{Palm tree} \\
        \bluebold{Image Manipulation: } \textit{Palm tree} is inserted into the image.
        \\[2pt]

    \begin{multicols}{2}

        \centering
        \includegraphics[width=0.6\columnwidth]{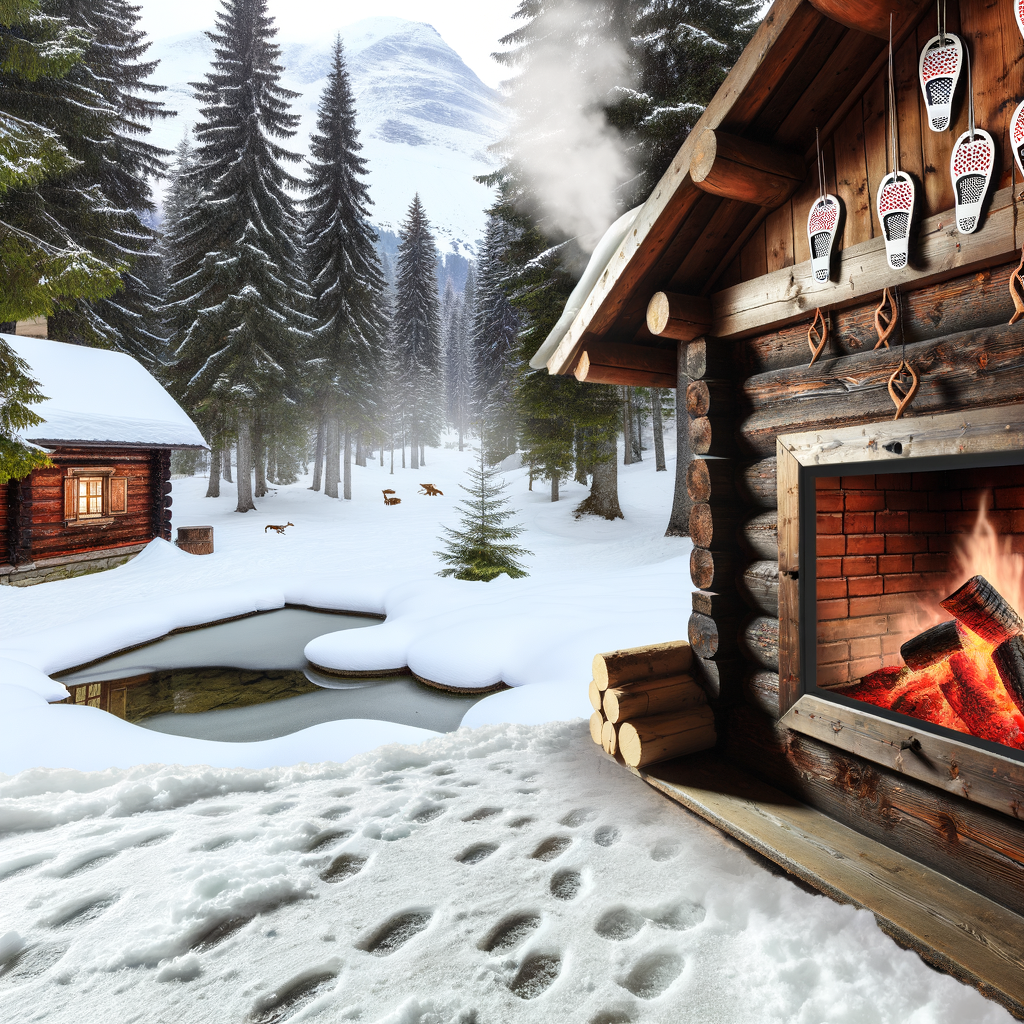}      \\
        \textit{\textbf{Original image}} \\        
        \columnbreak
        \centering
        \includegraphics[width=0.6\columnwidth]{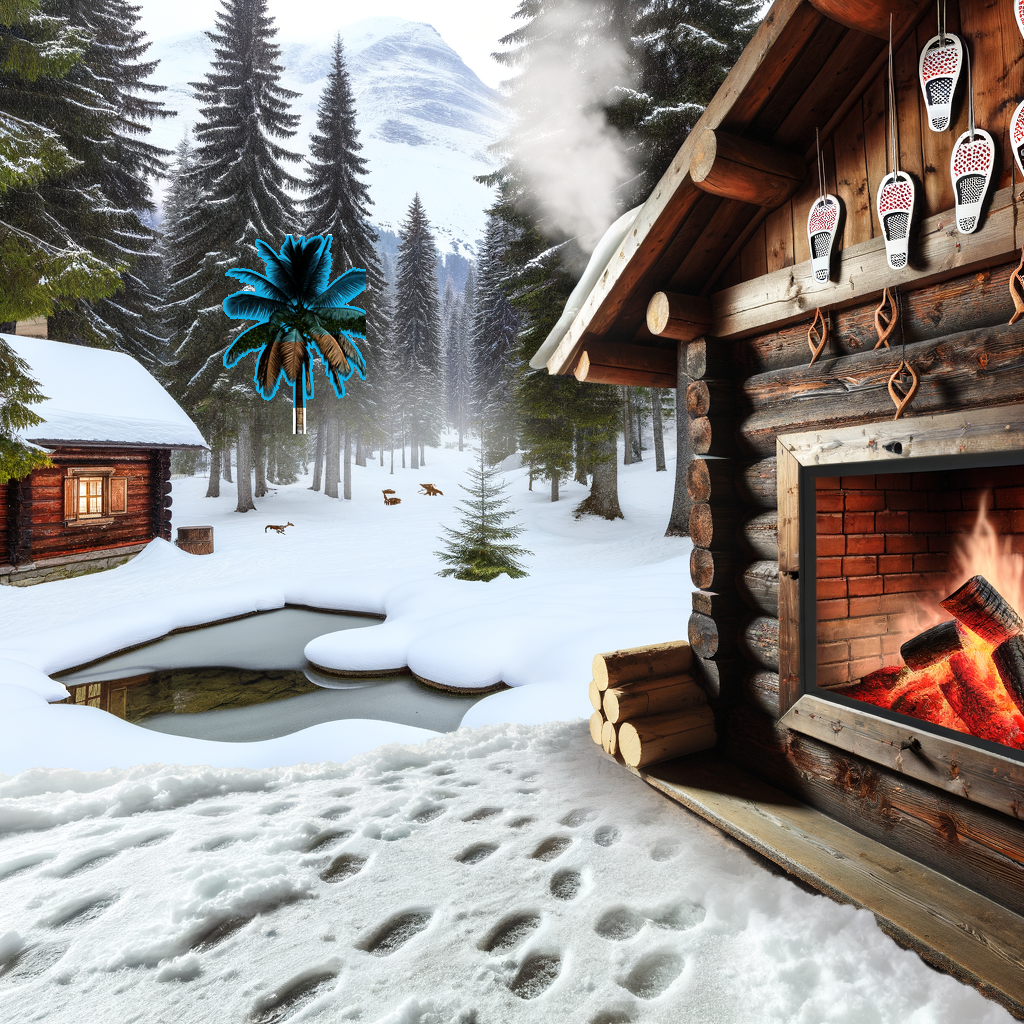} \\
        \textit{\textbf{Edited image}} \\
    \end{multicols}

        \raggedright
        \bluebold{Attack: } Adding new irrelevant object \textit{Palm tree} -- Existence \\
        \bluebold{Question: } Is there a/an Palm tree in this image? \\
        \bluebold{GPT-4V: } Yes, there is an image of a palm tree in this picture. It appears to have been artificially placed or edited into the snowy winter scene, as palm trees are not naturally found in such cold, snowy environments. The palm tree is located near the center of the image and is highlighted with a blue neon-like outline.\\
        \bluebold{LLaVA-1.6: } Yes, there is a palm tree in this image. It's quite unusual to see a palm tree in a snowy, mountainous setting, which makes the image quite surreal.\\
    
    \end{minipage}
};
\end{tikzpicture}%
\caption{\textbf{Failure Case: When LVLMs Detect the Edition.}}
\label{Failure Case: When LVLMs detect the edition}
\end{center}
\end{figure*}

\begin{figure*}
\begin{center}
\begin{tikzpicture}
\node [mybox,title=Failure Case: When Evaluation Model Fails] (box){%
    \begin{minipage}{2\columnwidth}
        \raggedright
        \bluebold{Scene Name: } \textit{Children petting goat} \\
        \bluebold{Detected Objects: } \textit{Black and white goat}, \textit{Floral summer dress}, \textit{Black domestic goat}, \textit{Illuminated digital keypad} \\
        \bluebold{Inserted Object: } \textit{Microwave} \\
        \bluebold{Image Manipulation: } \textit{Microwave} is inserted into the image.
        \\[2pt]

    \begin{multicols}{2}

        \centering
        \includegraphics[width=0.6\columnwidth]{failure_cases/eval/1/init.png}      \\
        \textit{\textbf{Original image}} \\        
        \columnbreak
        \centering
        \includegraphics[width=0.6\columnwidth]{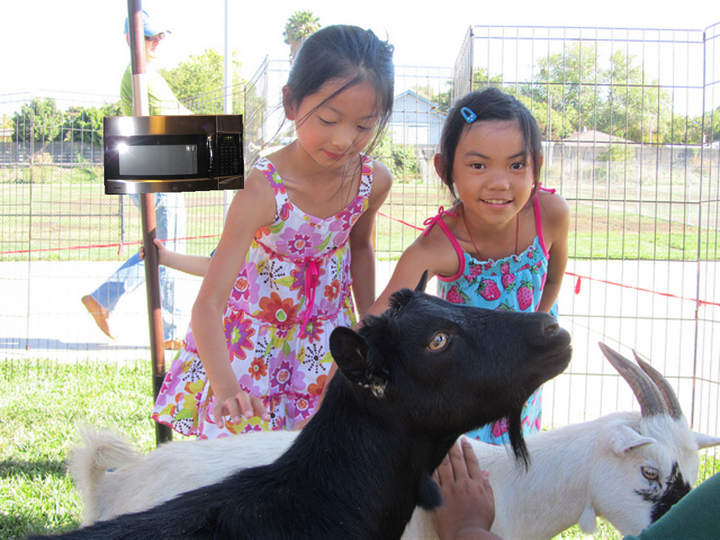} \\
        \textit{\textbf{Edited image}} \\
    \end{multicols}

        \raggedright
        \bluebold{Attack: } Adding new irrelevant object \textit{Microwave} -- Existence \\
        \bluebold{Question: } Is there a/an microwave in this image? \\
        \bluebold{GPT-4V: } No, there is not a microwave in the original context of the image. The photo shows two young girls interacting with goats in an outdoor setting, presumably at an animal pen or a small farm. The microwave appears to be an edited addition to the image. \\
        \bluebold{Ground Truth: } There is a microwave. \\
        \bluebold{Evaluation Result: } Not same. \\
        \quad \\ 
    \par\noindent\rule[15pt]{\textwidth}{1pt}\\[-30pt]
    \quad \\ 
    \raggedright
        \bluebold{Scene Name: } \textit{Cafe dessert table} \\
        \bluebold{Detected Objects: } \textit{Layered cream cake}, \textit{Stainless steel forks} \\
        \bluebold{Removed Object: } \textit{Glass} \\
        \bluebold{Image Manipulation: } One \textit{Glass} is removed from the image, while there's another \textit{Glass} is the edited image escaped from the object detection and removal model.
        \\[2pt]

    \begin{multicols}{2}

        \centering
        \includegraphics[width=0.6\columnwidth]{failure_cases/eval/2/init.png}      \\
        \textit{\textbf{Original image}} \\        
        \columnbreak
        \centering
        \includegraphics[width=0.6\columnwidth]{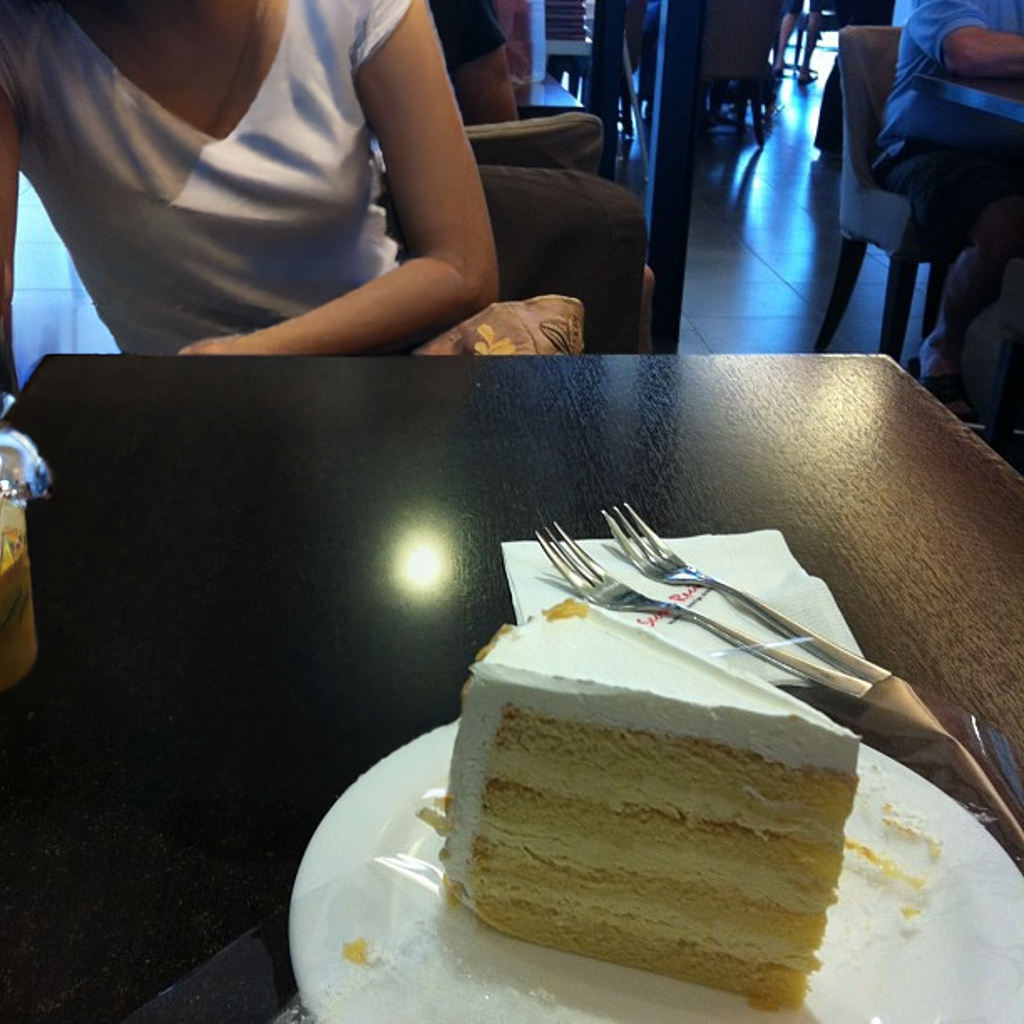} \\
        \textit{\textbf{Edited image}} \\
    \end{multicols}

        \raggedright
        \bluebold{Attack: } Remove an existing object \textit{Glass} -- Existence \\
        \bluebold{Question: } Is there a/an glass in this image? \\
        \bluebold{GPT-4V: } Yes, there is a glass in the image. It is located at the lower left portion of the picture, next to the piece of cake on the table.\\
        \bluebold{Ground Truth: } There is no glass.\\
        \bluebold{Evaluation Result: } Not same. \\
    
    \end{minipage}
};
\end{tikzpicture}%
\caption{\textbf{Failure Case: When Evaluation Model Fails.}}
\label{Failure Case: When Evaluation Model Fails}
\end{center}
\end{figure*}

We provide several cases for the failure situation of \ours{} we encountered in our experiment. Fig.~\ref{Failure Case: Non-Perceivable Objects Prompted} shows cases when a human could not understand the object being added. Fig.~\ref{Failure Case: When LVLMs detect the edition} shows cases when LVLMs detect the image manipulation we perform and point out in their answers. Fig.~\ref{Failure Case: When Evaluation Model Fails} shows cases when the evaluation model fails to provide the correct evaluation answer for the given VQA task. 

\input{showcases/insert_case_1}
\input{showcases/insert_case_2}
\input{showcases/corr_case_1}
\input{showcases/corr_case_2}
\input{showcases/removal_case_1}
\input{showcases/removal_case_2}


\end{document}